%% file: main.tex
\documentclass[11pt,letterpaper]{article}

\usepackage{preprint}

\input{preamble}
\usepackage[breaklinks,colorlinks,linkcolor=preprintlink,citecolor=preprintlink,urlcolor=preprintlink]{hyperref}

\preprintlogos{\raisebox{-0.5\height}{\includegraphics[height=5.6mm]{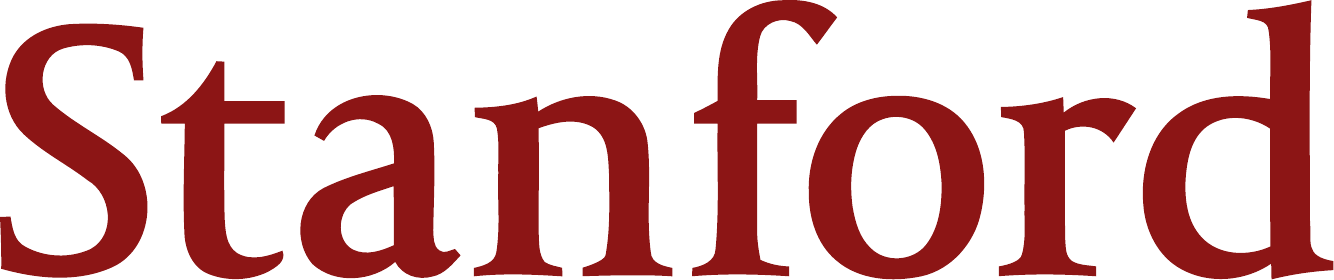}}\hspace{6.5mm}%
  \raisebox{-0.5\height}{\includegraphics[height=6mm]{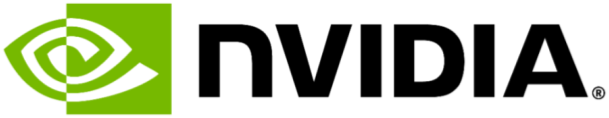}}}
\title{RoboVerify: A Self-Verification Framework for Agentic Robotics}
\preprintauthors{Yize Liu\textsuperscript{1,*}, Jacky Kwok\textsuperscript{1,*}, Yining Hong\textsuperscript{1}, Luis Pabon\textsuperscript{1}, Jadelynn Dao\textsuperscript{1}, Zhi Cao\textsuperscript{1}, Chongkai Gao\textsuperscript{1}, Yejin Choi\textsuperscript{1}, Azalia Mirhoseini\textsuperscript{1}, Leonidas Guibas\textsuperscript{1}, Marco Pavone\textsuperscript{2,1}}
\preprintaffiliations{\textsuperscript{1}Stanford University\quad \textsuperscript{2}NVIDIA}
\preprintnote{\textsuperscript{*}Equal contribution}
\preprintteaser{%
  \captionsetup{type=figure}%
  \includegraphics[width=\linewidth]{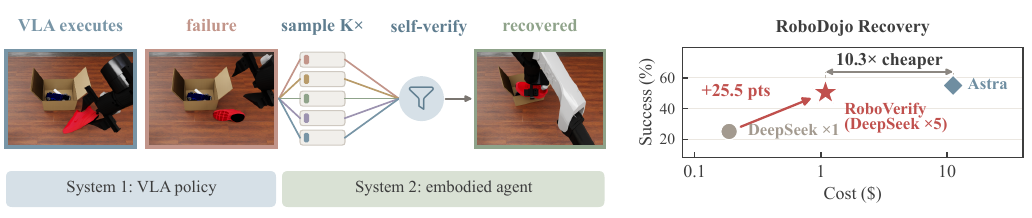}%
  \captionof{figure}{\textbf{A self-verification framework for LLM agents that control robots.} Left: a VLA policy executes a task; after it drops the shoe beside the box, an agent takes over and puts the shoe into the box. With \ours, the agent samples $K$ candidate actions per decision and verifies them to choose one. Right: recovery success on 220 RoboDojo episodes versus cost per episode. RoboVerify raises DeepSeek-V4.1-Flash close to GPT-6 Astra at 10.3$\times$ lower cost.}%
  \label{fig:teaser}}
\preprintwebsite{https://roboverify.github.io/}
\preprintcontact{yizeliu@stanford.edu}

\begin{document}
\input{sec/0_abstract}
\maketitle
\suppressfloats[t]   

\input{sec/1_intro}

\input{sec/2_related}
\input{sec/3_method}
\input{sec/4_benchmark}

\input{sec/5_experiments}
\input{sec/5b_real}
\input{sec/6_analysis}
\input{sec/7_conclusion}
\input{sec/8_acknowledgments}

{
    \small
    \bibliographystyle{ieeenat_fullname}
    \bibliography{main}
}

\input{sec/X_suppl}

\end{document}

%% file: preamble.tex
\makeatletter
\renewcommand\paragraph{\@startsection{paragraph}{4}{\z@}{1.2ex \@plus .3ex \@minus .2ex}{-1em}{\normalfont\normalsize\bfseries}}
\makeatother

\usepackage{multirow}
\usepackage{afterpage}
\usepackage{xspace}
\newcommand{\single}{\textsc{Single}\xspace}
\newcommand{\ours}{RoboVerify\xspace}
\newcommand{\astra}{\textsc{Astra}\xspace}
\newcommand{\gss}{grounded self-verification\xspace}
\newcommand{\bench}{TakeOver-520\xspace}
\newcommand{\suppBench}{\bench details}
\newcommand{\suppReal}{Real-robot setup}
\newcommand{\suppResults}{Full results}
\newcommand{\suppCost}{Cost and token accounting}
\newcommand{\suppAnalysis}{Analysis details}
\newcommand{\suppMore}{Additional analyses}
\providecommand{\eg}{e.g.\@\xspace}
\providecommand{\ie}{i.e.\@\xspace}
\providecommand{\vs}{vs.\@\xspace}

\usepackage{listings}

%% file: sec/0_abstract.tex
\begin{abstract}
Large language model (LLM) agents are a promising way to control robots, but frontier agents are costly.
A hierarchical design eases this: a vision-language-action (VLA) policy (System~1) executes most manipulation steps, and a tool-using agent (System~2) takes over only when it fails.
This cuts agent calls and latency, but each call stays costly: reliable frontier agents are expensive, while cheaper models fail far more often.
We find that this gap is mostly one of \emph{reliability}, not \emph{capability}: a cheaper model can often produce the right recovery action, but a single sample is unstable.
We introduce \ours, a model-agnostic, training-free self-verification framework that samples several candidate actions per decision and lets the same model verify them and choose one.
At its core is \gss: the verifier judges candidates against code-computed geometric evidence pooled across samples, not their explanations.
To evaluate failure-time takeover, we build \bench, a benchmark of 520 real $\pi_{0.5}$ failures in three simulation suites (RoboDojo, RoboCasa, PolaRiS).
On \bench, \ours closes 83\% of the gap in recovery success between DeepSeek-V4.1-Flash and the frontier agent GPT-6 Astra, at 7.4--10.3$\times$ lower cost.
It also transfers to complete tasks: on a real robot, it raises success from 20.8\% to 75.0\%, close to the frontier agent.
These results suggest that, on some tasks, cheaper models can replace frontier models in robot control at much lower cost.
\end{abstract}

%% file: sec/1_intro.tex
\section{Introduction}
\label{sec:intro}

Embodied agents built on large language models (LLMs) have become capable robot operators.
Given camera views and perception and geometry tools, such an agent can inspect a scene, reason about it, and plan motions step by step~\cite{liang2023cap,huang2023voxposer,fu2026capx,yang2026ora}.
Running it on a frontier model such as GPT-6 Astra~\cite{openai2026astra}, however, is expensive.
Every decision consumes a long multimodal context, and even a short manipulation episode takes many rounds of reasoning and decisions.

A natural way to contain this cost is to pair a fast learned policy with an agent (\cref{fig:teaser}).
A vision-language-action (VLA) policy such as $\pi_{0.5}$~\cite{black2025pi05} (``System~1'') handles routine execution, and the agent (``System~2'') takes over only when the policy fails or is about to fail, to diagnose the situation and recover from it.
This division of labour limits how often the agent is called, but not what each call costs: a takeover is itself a closed loop of many observe--decide--act rounds, and with a frontier agent every takeover remains expensive.
We take this framework as given and ask whether the takeover can be carried out by a much cheaper model.

Simply substituting a cheaper model does not work: it recovers far fewer failures than the frontier agent.
The usual explanation is that the cheaper model lacks the necessary spatial understanding and motion planning.
We argue instead that, in robot control, much of the gap lies elsewhere.

Robot control differs from typical language and coding tasks in that the agent cannot plan to the end.
The outcome of each motion depends on contacts and dynamics that the agent cannot predict in advance, so it must act in short chunks, observe the result, and decide again.
Task success therefore requires every decision along the way to be sound.
A single faulty decision, such as closing the gripper on nothing or knocking a block off a tower, can end the episode.
Here, a small difference in how reliably each decision is made turns into a large difference in task success.

Our key insight is that this gap is largely one of \emph{reliability} rather than \emph{capability}.
A cheaper model often understands what a recovery requires and can produce the right action, but it does not do so consistently.
Asked to decide several times from the same state, it may choose different actions, and occasionally even different objects to act on.
A frontier agent is not necessarily able to do more; it is simply more likely to get each decision right, and over the many decisions of a closed-loop recovery this advantage adds up.

To make each decision of a cheaper model more reliable, we build on a familiar observation in agent research: judging is easier than generating~\cite{cobbe2021verifiers,song2024mindgap}.
Picking a correct action from a few concrete candidates is easier than producing one in a single attempt.
If several candidate actions are drawn at every decision, it becomes likely that a good one is among them, and what remains is to pick it out.
Self-verification without external feedback, however, is known to be unreliable~\cite{huang2024cannot}: a model that judges its own outputs by their explanations tends to be persuaded by them.
For self-verification to help, it has to rest on evidence that the candidates cannot talk their way around.

Building on this, we introduce \ours, a model-agnostic, training-free self-verification framework that amplifies the reliability of an embodied agent at test time.
At every decision, one model generates several candidate actions from the same state (\emph{candidate sampling}) and then verifies them to choose one to execute (\emph{\gss}).
The verifier treats each candidate's explanation as an unverified claim and judges the candidates against a code-computed geometric check table and the perception measurements pooled from all candidates (\cref{sec:method}).
\ours needs no separate verifier and no training.

To evaluate failure-time takeover, we also build \bench, a benchmark of real failures of a pretrained $\pi_{0.5}$ policy in three simulation suites.
The agent must complete the interrupted subgoal without being told what it is (\cref{sec:benchmark}).
On \bench, we compare a cheaper model with and without \ours against a frontier agent (\cref{sec:experiments}) and examine where the differences between them come from (\cref{sec:analysis}).
With \ours, the cheaper model recovers substantially more failures and closes most of its gap to the frontier agent, at a small fraction of the frontier agent's cost.
Zero-shot experiments on atomic tasks with a real robot show the same pattern (\cref{sec:real}).

%% file: sec/2_related.tex
\section{Related Work}
\label{sec:related}

\paragraph{VLA policies and their failures.}
Generalist VLA policies map images and language directly to actions~\cite{brohan2023rt2,kim2024openvla,black2024pi0,black2025pi05}.
They achieve broad coverage but degrade under small perturbations, and their failures are hard to correct without retraining~\cite{wang2024vlatest,zhou2025liberopro,fei2025liberoplus}.
Prior work detects such failures at run time~\cite{agia2024sentinel,gu2025safe} or explains and corrects them, often with models trained on failure data~\cite{liu2023reflect,shi2024yay,dai2025racer,duan2025aha,lin2025failsafe}.
In contrast, \ours leaves the policy unchanged and recovers with an off-the-shelf model, without failure data or training.

\paragraph{Foundation-model agents for manipulation.}
Language and vision-language models can write robot code, compose perception and planning tools, and reason about scenes~\cite{liang2023cap,huang2023voxposer,ahn2022saycan,huang2024rekep,fu2026capx}.
Such agents are strong at diagnosis and spatial reasoning but too slow and costly for every control step, which motivates hierarchical vision-language-action (Hi-VLA) systems in which a high-level model directs a fast low-level policy~\cite{shi2025hirobot,bjorck2025gr00t,song2025hume,hu2026hivla}.
Our harness follows this design~\cite{yang2026ora} and calls the agent only after a failure.

\paragraph{Sampling, verification and test-time compute.}
Drawing several samples and verifying or voting on them improves language-model reasoning and code generation~\cite{li2022alphacode,wang2023selfconsistency,cobbe2021verifiers,lightman2024verify,brown2024monkeys,snell2024scaling}.
Test-time sampling has also been applied to VLA actions, with learned verifiers~\cite{kwok2025robomonkey,nakamoto2024vgps,wu2025forewarn,kwok2026cover} or the policy's own confidence~\cite{jang2026mgselect}.
In contrast, \ours uses no trained verifier: the generator verifies its own samples, grounded in code-computed geometry and cross-sample measurements.
Aggregating cheaper language models can surpass a frontier one on language benchmarks~\cite{wang2025moa}; we ask whether a fixed, small number of samples makes a cheaper robot agent as reliable as a frontier one.
Self-correction without external feedback is known to be unreliable~\cite{huang2024cannot,stechly2025selfverification}, whereas critiques grounded in external tools help~\cite{gou2024critic}; our results suggest that grounding in measured evidence is what makes self-verification useful here.

%% file: sec/3_method.tex
\section{Method}
\label{sec:method}
Our goal is to make a cheaper model recover from the failures of a VLA policy as reliably as a frontier agent.

\subsection{Preliminaries}
\label{sec:setting}

\paragraph{Per-round reliability compounds.}
A recovery proceeds in \emph{rounds} of observing, deciding, and acting.
We call a round \emph{reliable} if its executed action keeps the recovery on a path to completing the interrupted subgoal: it neither makes an irrecoverable mistake nor stalls.
Let $q$ be the probability that a round is reliable.
Treating rounds as independent, a recovery that needs $n$ rounds succeeds with probability
\begin{equation}
  P_{\text{succ}}(n) = q^{\,n}, \qquad
  \frac{P'_{\text{succ}}(n)}{P_{\text{succ}}(n)} = \Bigl(\frac{q'}{q}\Bigr)^{n},
  \label{eq:compound}
\end{equation}
so raising it from $q$ to $q'$ multiplies task success by a factor exponential in the horizon $n$.
Long recoveries are where an unreliable agent fails and where a more reliable one gains the most.
Software agents show the same constant failure rate per unit of task length~\cite{kwa2025measuring,ord2025halflife}.
We estimate this equivalent per-round reliability $q$ in \cref{sec:analysis}.

\paragraph{Sampling and verification.}
Suppose a single sample from the agent is reliable with probability $p$, so a single-sample agent has $q_1 = p$.
Draw $K$ independent candidates from the same state and let a verifier choose one of them to execute.
At least one candidate is reliable with probability $c_K = 1-(1-p)^K$, and if the verifier picks a reliable candidate with probability $\sigma$ whenever one exists, then
\begin{equation}
  q_K = \sigma \, c_K = \sigma \bigl[1-(1-p)^K\bigr].
  \label{eq:qk}
\end{equation}
Sampling raises the \emph{coverage} $c_K$, which approaches 1 as $K$ grows; verification decides how much of it is realised.
A random pick gives $\sigma_0 = p/c_K$ and therefore $q_K = p$: sampling alone does not help.
\Cref{eq:qk} thus exceeds the single-sample reliability exactly when the verifier does better than chance, $\sigma > \sigma_0$, and with a perfect verifier it reaches $c_K$.
Combining \cref{eq:compound,eq:qk}, the gain in task success is $(\sigma c_K / p)^n$.
\ours is designed to make $\sigma$ high without a second model (\cref{sec:gss}).

\begin{figure}[t]
  \centering
  \includegraphics[width=\linewidth]{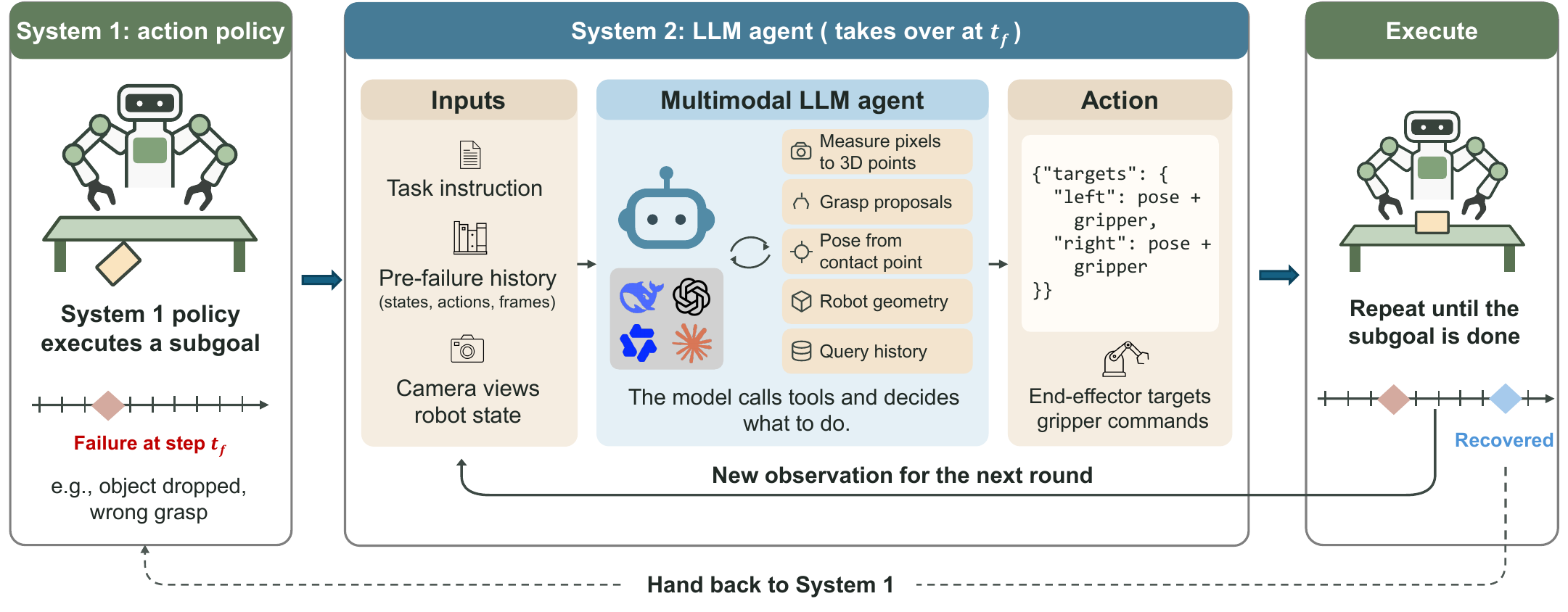}
  \caption{\textbf{Failure-time recovery agent.} At each round the agent observes the scene, calls perception and geometry tools, and proposes an action. The verified action is executed for several control steps and the loop repeats.}
  \label{fig:arch}
\end{figure}

\subsection{Tool-using recovery harness}
\label{sec:harness}
\Cref{fig:arch} shows our harness for recovering from the failures of a System~1 policy.
When the low-level policy has failed or is about to fail, a higher-level System~2 agent takes over: a multimodal language model with tool access, such as DeepSeek-V4.1-Flash~\cite{deepseek2026v41flash} or GPT-6 Astra~\cite{openai2026astra}.
In each round, the agent receives the task goal, the history, and the current state.
It reasons on its own, decides which tools to call to measure the scene, and finally outputs a structured action.
We build the harness for failure-time takeover and adapt its tool-calling interface and camera-geometry code from ORA~\cite{yang2026ora}.
The tools (i) turn image pixels and depth into 3D points and the vectors between them, (ii) propose grasp poses for an image region or point~\cite{sundermeyer2021contact}, (iii) convert a chosen contact point and approach direction into an end-effector pose, (iv) report the gripper geometry and the current approach and jaw axes, (v) compute paired targets when both arms carry one object, and (vi) read the execution history.
The controller executes the action returned at the end of a round and then returns the new observation.

\subsection{\ours: Candidate Sampling and Grounded Self-Verification}
\label{sec:gss}

\begin{figure}[t]
  \centering
  \includegraphics[width=\linewidth]{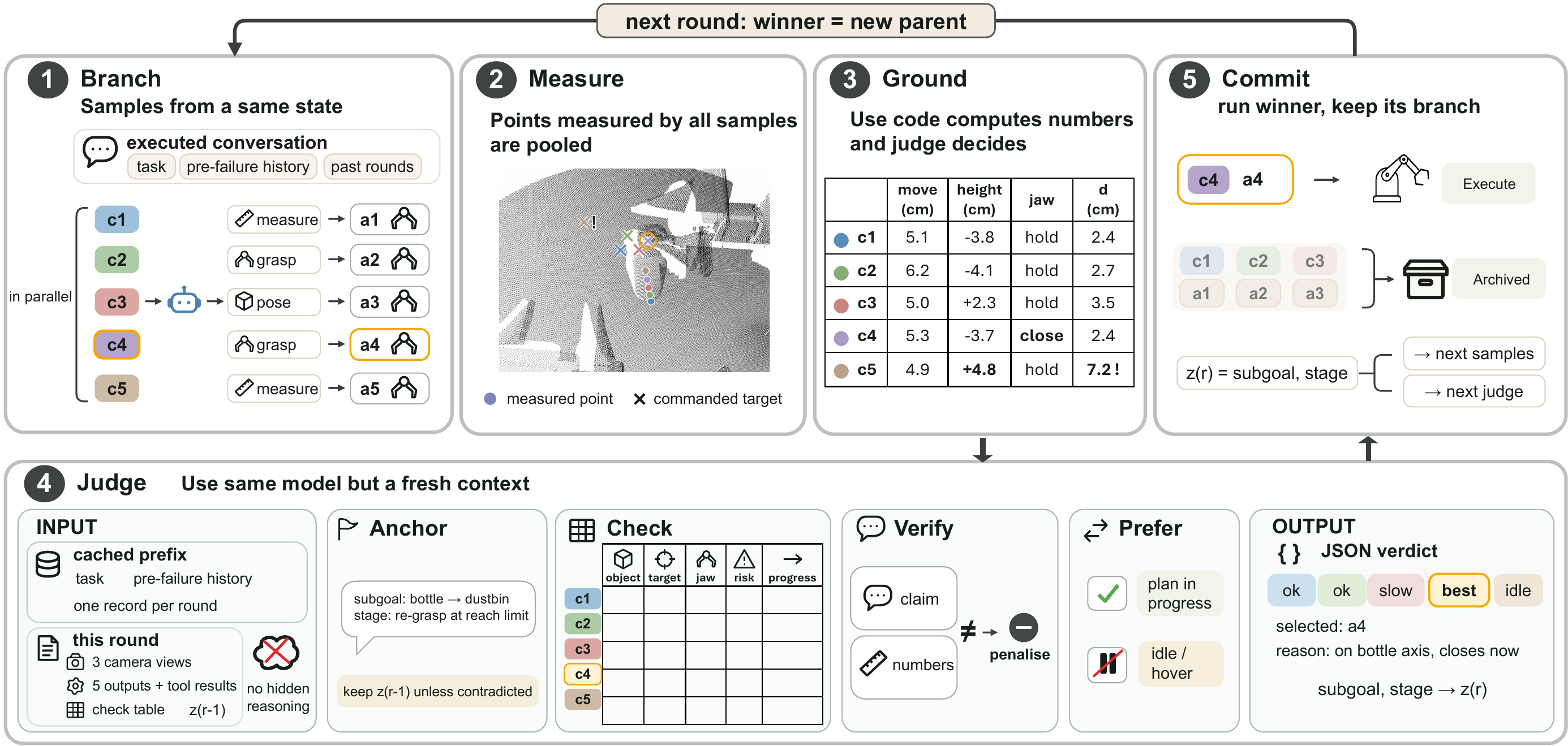}
  \caption{\textbf{One round of \ours.} (1) \emph{Branch}: $K$ candidates are sampled from the executed state. (2) \emph{Measure}: the 3D points measured by all candidates are pooled. (3) \emph{Ground}: code computes the geometric check table; the numbers are from one logged round. (4) \emph{Judge}: the verifier, the same model in a fresh context, anchors the subgoal, checks the candidates against the table, and returns its choice. (5) \emph{Commit}: the chosen action is executed unchanged, and its branch and the note $z_r$ pass to the next round.}
  \label{fig:selection}
\end{figure}

Part of the gap between models lies in the reliability of their outputs, and \ours addresses it with test-time scaling: \emph{candidate sampling} followed by \emph{\gss} (\cref{fig:selection}).
More test-time compute alone is not enough; the tools and the design of verification are what bring out the model's full potential.
The core idea is that a model that cannot directly produce reliable answers should rest both its generation and its verification on deterministic tool results, which makes its judgments more accurate.
Concretely, at every round we draw $K{=}5$ candidates in parallel from the same model and the same conversation state (step~1).
Each candidate is an independent tool-using generation, parsed into an executable action before verification.
A verifier then checks the candidates and picks one, which is executed without modification.
The verifier is the same model as the generator, so this framework needs no separate verifier model and no additional training, and could apply to any model.

\begin{figure}[t]
  \centering
  \includegraphics[width=\linewidth]{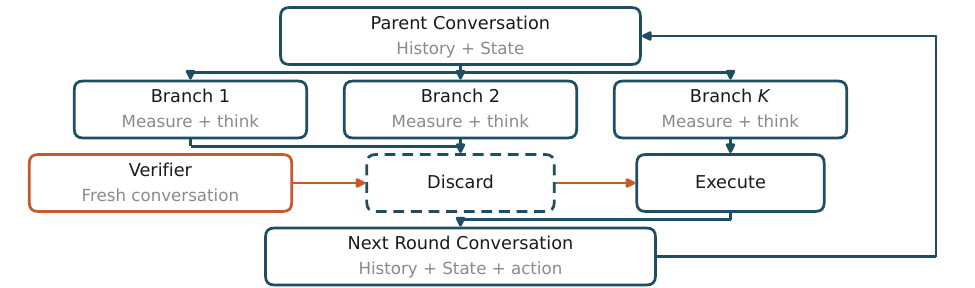}
  \caption{\textbf{Context inheritance across rounds.} The $K$ branches copy the parent conversation (history and state), so its prefix is shared and cached; each branch measures and reasons on its own. The verifier, the same model in a fresh conversation, chooses one branch; it is executed and becomes the parent of the next round, and the others are discarded.}
  \label{fig:context}
\end{figure}

\paragraph{Context inheritance.}
Every candidate inherits the whole recovery so far, and only the verified one is carried forward (\cref{fig:context}).
Each round starts from a \emph{parent} conversation that holds the task, the pre-failure history, and every executed round so far, including the executed candidate's tool calls and reasoning.
The $K$ candidates are $K$ branches copied from this parent.
Each branch continues on its own: it measures the scene with its own tool calls, reasons about the result, and ends with one action, without seeing its siblings.
Because every branch begins with the same parent verbatim, this shared prefix is served from the provider's prompt cache; across the three suites, 98--99\% of the input tokens of \ours are cache hits.
After verification, the winner's conversation becomes the parent of the next round (step~5) once its action has been executed and the new observation has arrived.
The other $K{-}1$ branches are archived and never enter any later context.
The verifier inherits none of these conversations.
It opens a fresh context every round from the task, the pre-failure history, and the current state, plus one short record per executed round: the state at decision time, the executed candidate's output, and the subgoal note.
Its design follows five principles, as the steps of \cref{fig:selection}.

\paragraph{1. Cross-sample perception.}
Code pools the 3D points that \emph{all} candidates measured (step~2).
A candidate that misperceives an object thus disagrees with its siblings' measurements, which a single sample cannot reveal.

\paragraph{2. Evidence over explanation.}
The verifier judges each candidate against numbers computed by code, which an explanation cannot talk its way around.
For every candidate and arm, code computes a \emph{geometric check table} (step~3) from public measurements: the commanded gripper-pad target and current pad position, the displacement and its vertical component (\emph{move}, \emph{height}), the commanded and measured gripper opening and timing (\emph{jaw}), and the three nearest pooled points with their distances ($d$).
The verifier must penalise any candidate whose numbers contradict its claims (step~4, \emph{Verify}).
The table only reports quantities; it never decides pass or fail.

\paragraph{3. Role separation.}
The verifier (step~4) sees what each candidate proposed and what its tools measured, but not the reasoning behind it, and its prompt states that candidate explanations are unverified claims.
It therefore judges the actions without adopting, or being swayed by, any candidate's line of thought.
It may only choose, never rewrite, so the executed action is always one that passed the action parser.

\paragraph{4. Intent anchoring.}
Before judging, the verifier states the interrupted subgoal and the current stage from the task, the history, and the images (step~4, \emph{Anchor}).
This note $z_r$ is passed to the next round, and the subgoal is kept unless the evidence clearly contradicts it.
Since the agent is never told the subgoal, this guards against switching objects or goals.

\paragraph{5. Progress over hesitation.}
The verifier prefers the candidate that continues a plan in progress unless it has visibly failed, and avoids candidates that only hover or re-observe when progress is possible (step~4, \emph{Prefer}).
A stalled round spends control budget without advancing the subgoal.

\paragraph{Verifier protocol and output.}
The verifier reasons with extended thinking, and its prompt fixes the order of step~4: \emph{Anchor}, \emph{Check}, \emph{Verify}, \emph{Prefer}.
In \emph{Check}, it rates every candidate on five criteria: (i) whether it acts on the right object, (ii) whether its commanded targets match the measured object and gripper poses, (iii) whether its gripper timing is right, (iv) whether it risks a collision or a drop, and (v) whether it makes concrete progress within the next 10 control steps.
Contact with other objects counts only if it prevents completing the subgoal.
The verifier answers in JSON with the subgoal, the stage, a one-line verdict per candidate, the index of the chosen candidate, and a reason.
A reply that cannot be parsed is returned to the verifier with the parsing error, up to three calls per round; in our runs the fallback to a default candidate was never needed.

\paragraph{Cost.}
Verification adds one model call per round, with an append-only prompt that is mostly served from the prompt cache.
It accounts for 15--16\% of the cost of \ours (\cref{sec:supp:cost}).

%% file: sec/4_benchmark.tex
\section{\bench: A Benchmark of Real Policy Failures}
\label{sec:benchmark}

\begin{table}[t]
  \centering
  \small
  \setlength{\tabcolsep}{4pt}
  \begin{tabular}{@{}lcccc@{}}
    \toprule
    Suite & Tasks & Episodes & Budget & Success check \\
    \midrule
    RoboDojo~\cite{chen2026robodojo} & 11 & 220 & 300 & native, subgoal \\
    RoboCasa~\cite{nasiriany2024robocasa} & 18 & 180 & 400 & native \\
    PolaRiS~\cite{jain2025polaris} & 6 & 120 & 300 & native, subgoal \\
    \midrule
    Total & 35 & 520 & & \\
    \bottomrule
  \end{tabular}
  \caption{\textbf{Composition of \bench.} Every episode starts from an exactly replayed real failure of the suite's released $\pi_{0.5}$ policy. Budget is in control steps after the failure.}
  \label{tab:benchmark}
\end{table}

\bench contains 520 episodes from 35 tasks across three simulation suites (\cref{tab:benchmark}).

\paragraph{Collecting failures.}
We run $\pi_{0.5}$~\cite{black2025pi05} on the native tasks of each simulation suite, using the checkpoint that its authors fine-tuned and released.
These are the RoboDojo baseline~\cite{chen2026robodojo,xpolicylab2026}, the RoboCasa365 multitask policy~\cite{nasiriany2026robocasa365}, and the $\pi_{0.5}$-DROID policy released with PolaRiS~\cite{jain2025polaris,khazatsky2024droid} (checkpoints, task lists, and predicates: \cref{sec:supp:benchmark}).
Failure events are proposed automatically from the recorded object and gripper states and confirmed by human review.
Each confirmed failure yields an episode of one of two kinds, which differ in where the agent takes over.
In a \emph{failure-state} episode, the agent starts at the first control step at which the failure is visible and must recover from it.
In a \emph{pre-failure} episode, the failure is one the agent could not undo, such as an object pushed out of reach; the agent instead takes over at the last normal state before the policy's irrecoverable action and completes the subgoal from there.
Success means completing the interrupted subgoal: \eg, placing a dropped block back on the tower or one more item into the box.
We check it with the suite's native task predicates restricted to that subgoal.
Violations of native task constraints (\eg, lifting a block that must only be pushed, or pressing a button too often) count as failures.

\paragraph{Suites.}
RoboDojo-220 contains 220 episodes from 11 bimanual tabletop tasks, 20 each: six from the original suite, such as building a tower, packing a box, and hanging mugs, and five further tasks, such as pushing a T and pressing buttons by number.
RoboCasa-180 contains 180 episodes from 18 atomic kitchen tasks, 10 each, covering doors, drawers, lids, appliances, and pick-and-place between counters, cabinets, sinks, and appliances.
PolaRiS-120 contains 120 episodes from six tabletop tasks, 20 each, such as stacking blocks, bussing food into a bowl, and scrubbing a pan.

%% file: sec/5_experiments.tex
\section{Experiments}
\label{sec:experiments}

We ask how much of the gap to a frontier agent \ours closes, at what cost (\cref{sec:main}), whether this holds on a real robot (\cref{sec:real}), and where the gain comes from (\cref{sec:analysis}).

\paragraph{Methods and metrics.}
Three recovery agents share the harness, prompts, controller, and budget of \cref{sec:method} and differ only in the model and the number of candidates per round:
\begin{itemize}[leftmargin=*,itemsep=1pt,topsep=2pt]
  \item \single: DeepSeek-V4.1-Flash~\cite{deepseek2026v41flash} (snapshot 1017), one candidate per round.
  \item \ours: the same model with $K{=}5$ candidates per round and \gss.
  \item \astra: a frontier proprietary agent model, GPT-6 Astra~\cite{openai2026astra} at high reasoning effort, one candidate per round.
\end{itemize}
We report \emph{recovery success}, the fraction of episodes that complete the interrupted subgoal within budget, and \emph{cost} per episode: the API cost of all model calls, computed from logged token counts at list prices (USD per million cached input, uncached input, and output tokens): 0.003, 0.15, and 0.60 for DeepSeek-V4.1-Flash, and 1, 10, and 50 for \astra, whose requests above 272K input tokens are charged 2$\times$ for input and 1.5$\times$ for output.

\paragraph{Simulation setup.}
RoboDojo simulates two ARX X5 arms in Isaac Sim; RoboCasa simulates a Franka Panda on a mobile base in MuJoCo kitchen scenes; PolaRiS simulates a fixed-base Franka Panda with a Robotiq gripper at 15\,Hz in scenes reconstructed from real environments with Gaussian splatting.
In every suite the agent receives only what a real robot provides: calibrated RGB-D camera views (three on RoboDojo and RoboCasa, two on PolaRiS) and the robot's proprioception (end-effector poses and gripper openings).
The commanded end-effector targets are tracked by each suite's own controller (an operational-space controller on RoboCasa, where the agent can also move the base and torso).
Per-suite details are in \cref{sec:supp:benchmark}.

\subsection{Main results}
\label{sec:main}
\begin{figure}[t]
  \centering
  \includegraphics[width=\linewidth]{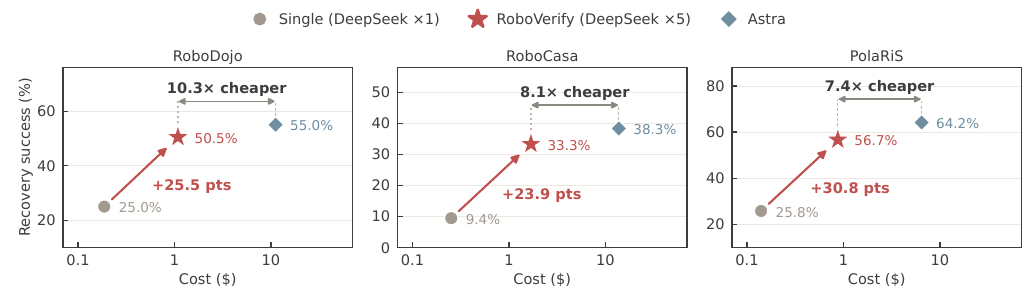}
  \caption{\textbf{Recovery success versus cost per episode} on TakeOver-520 (cost on a log scale). The red arrow marks the gain of \ours over \single; the grey double arrow marks how much cheaper \ours is than \astra. \ours closes 80--85\% of the gap to \astra at 7.4--10.3$\times$ lower cost. Per-suite numbers, cost per success, and tokens are in \cref{sec:supp:results}.}
  \label{fig:cost}
\end{figure}

\begin{figure}[t]
  \centering
  \includegraphics[width=\linewidth]{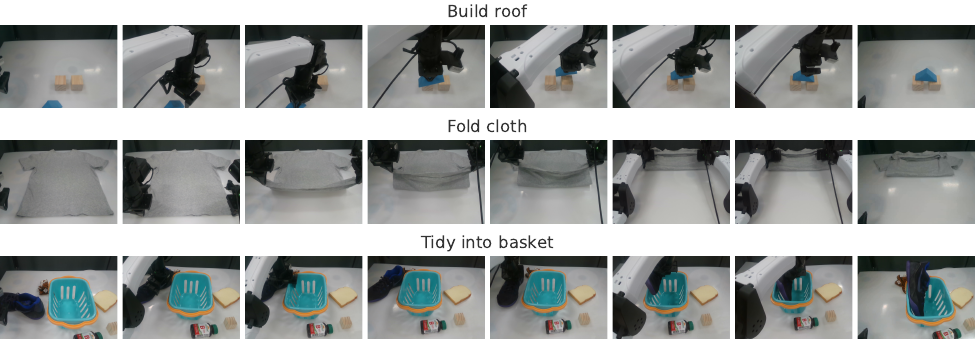}
  \caption{\textbf{Real-robot tasks.} One successful run per task, seen from the fixed overhead camera, from the start (left) to the completed task (right).}
  \label{fig:real}
\end{figure}

\paragraph{Sampling and verification close most of the gap to the frontier agent.}
Over all 520 episodes, \ours raises recovery success from 19.8\% to 46.0\%, against 51.4\% for \astra, closing 83\% of the gap (\cref{fig:cost}; per-suite numbers and per-task success: \cref{sec:supp:results}).
On RoboDojo, it raises recovery success from 25.0\% to 50.5\%, against 55.0\% for \astra, closing 85\% of the gap.
On RoboCasa, it raises recovery success from 9.4\% to 33.3\% (\astra: 38.3\%), and on PolaRiS from 25.8\% to 56.7\% (\astra: 64.2\%), closing 83\% and 80\% of the gap.
The gain is consistent across three simulators, single-arm and bimanual robots, and tasks ranging from block stacking to articulated kitchen appliances: on every suite, \ours at least doubles the success of \single (2.0$\times$ on RoboDojo, 3.5$\times$ on RoboCasa, 2.2$\times$ on PolaRiS).
The model, tools, and controller are the same as for \single, so this gain comes from how each decision is made, not from new capabilities.
This supports our view that the gap to the frontier agent is largely one of reliability.

\paragraph{At a fraction of the cost.}
\ours is 7.4--10.3$\times$ cheaper than \astra, although it draws five candidates and makes one verification call per round.
It costs \$1.09 per episode on RoboDojo against \$11.18 for \astra (10.3$\times$ less), \$1.69 against \$13.76 on RoboCasa (8.1$\times$), and \$0.87 against \$6.45 on PolaRiS (7.4$\times$).
It processes 15--25 times more tokens than \astra, but at a far lower price per token and with 98--99\% of its input served from the prompt cache (\cref{sec:supp:results,sec:supp:cost}).
Spending test-time compute on a cheaper model is therefore a much cheaper route to reliable recovery than a frontier model.

\paragraph{\ours solves largely the same failures as the frontier agent.}
Episode by episode, most of the failures that \ours recovers are also recovered by \astra.
This is what we expect if the cheaper model already has the capability and lacks only reliability.
\ours also keeps most of what \single achieves and adds many failures that \single does not recover, while very few are recovered by \single alone.
Sampling and verification thus extend the set of failures that the model recovers (per-episode breakdown: \cref{sec:supp:results}).

%% file: sec/5b_real.tex
\subsection{Real-robot experiments}
\label{sec:real}

\paragraph{Setup.}
We evaluate on a bimanual platform with two 6-DoF I2RT YAM arms, each with a force-limited parallel-jaw gripper.
The agent observes three RGB cameras (a fixed overhead camera and one on each wrist) together with the measured joint states, end-effector poses, and gripper openings.
As in simulation, it outputs an end-effector target pose and a gripper command per arm; a server on the robot host converts the targets into joint targets by inverse kinematics and tracks them at 25\,Hz.
A round spans 20 control steps (0.8\,s): the target is interpolated under a per-step limit, and the arm then settles.
An episode has at most 300 control steps, \ie 12\,s of commanded motion.
All three methods use the same server, motion limits, task instructions, and reasoning effort (details: \cref{sec:supp:real}).

\paragraph{Tasks.}
On the real robot, we test whether \ours extends from recovering failures to completing whole tasks.
Hence, no $\pi_{0.5}$ policy runs first: the agent performs three atomic tasks that we design from start to finish, each testing a different ability (\cref{fig:real}).
\emph{Build roof} asks the agent to place a roof block on top of two cubes and tests placement accuracy.
\emph{Fold cloth} asks it to fold a T-shirt lying on the table and tests two-arm coordination.
\emph{Tidy into basket} asks it to find the object named in the instruction (a wooden block, a cup, a long spoon, a pill bottle, scissors, a shoe, or a stuffed toy) among several objects and put it into a basket.
It tests object recognition and grasp judgement.

\begin{figure}[t]
  \centering
  \includegraphics[width=0.65\linewidth]{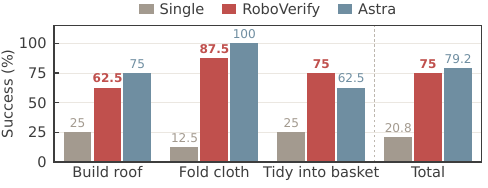}
  \caption{\textbf{\ours approaches the frontier agent on the real robot.} Success rate (\%) per task and method.}
  \label{fig:realres}
\end{figure}

\paragraph{Results.}
On the real robot, \ours raises the success rate from 20.8\% with a single sample to 75.0\%, close to the frontier agent's 79.2\% (\cref{fig:realres}).
It improves every task, most of all cloth folding (from 12.5\% to 87.5\%).
The pattern matches simulation: sampling and grounded verification turn an unreliable cheaper model into one that succeeds about as often as the frontier agent.

%% file: sec/6_analysis.tex
\section{Why Does \ours Work?}
\label{sec:analysis}

\cref{sec:setting} splits the success of \ours into three links: sampling must make the right action \emph{available} (the coverage $c_K$ in \cref{eq:qk}), verification must \emph{find} it more often than a single draw does ($\sigma>\sigma_0$), and the resulting gain per round must \emph{compound} over the recovery (\cref{eq:compound}).
We examine each link in turn; details are in \cref{sec:supp:analysis_details}.

\begin{figure}[t]
  \centering
  \includegraphics[width=0.6\linewidth]{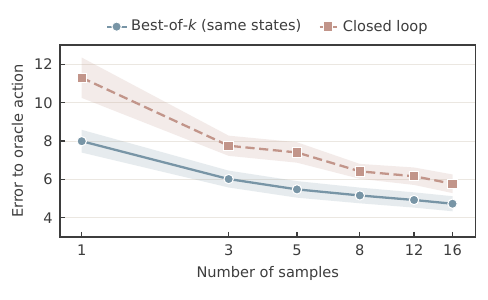}
  \caption{\textbf{Test-time scaling: more samples cover a better action.} Distance between the gripper target of the best of $k$ candidates and the correct next action, given by an oracle, on 30 episodes ($\pm 1$ standard error). Best-of-$k$: on the same states; closed loop: a separate run for each $k$, executing the best candidate.}
  \label{fig:scaling}
\end{figure}

\paragraph{Sampling makes the right action available.}
To measure what the samples contain, independently of which one is executed, we use failures whose correct recovery is unique.
A scripted oracle that reads the true simulator state then knows the correct next action.
In every round the model draws $k$ candidates, and the oracle, not the model, executes the one closest to it.
The error falls steadily as $k$ grows (\cref{fig:scaling}): five samples, the number \ours uses, already remove most of the error of a single sample (8.0\,cm to 5.5\,cm on the same states), and the curve has not flattened at sixteen.
The same holds in closed loop, where each run visits its own states.
The correct action is thus often among the samples even when a single sample misses it: the coverage $c_K$ rises quickly with $K$.
We read this as the signature of a test-time scaling law: more test-time compute raises the chance that the correct action is among the candidates.

\paragraph{A single sample is a gamble.}
The main runs show the same instability without an oracle.
Since \ours draws five candidates per round from the same state, its logs show how much the model's answer to one decision varies.
In 44.2\% of rounds the five candidates differ substantively (different targets, gripper actions, arms, or objects), and in 12.8\% they do not agree on what to do at all (\cref{sec:supp:analyses}).
Which action a single-sample agent executes is thus partly a matter of chance.
\ours replaces this chance with a choice grounded in measured evidence (\cref{sec:gss}); in a quarter of the rounds the verifier executes a substantively different action from the one a single-sample agent would take (\cref{sec:supp:analyses}).

\begin{figure}[t]
  \centering
  \includegraphics[width=0.6\linewidth]{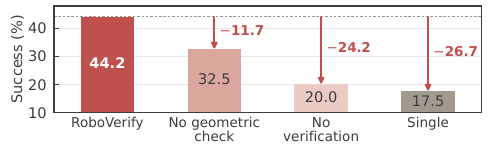}
  \caption{\textbf{Both verification and the geometric check table matter.} End-to-end ablation on TakeOver-520-RoboDojo episodes; arrows mark the gap to \ours. No verification executes a random valid candidate. }
  \label{fig:ablation}
\end{figure}

\paragraph{Verification finds the right action.}
By \cref{eq:qk}, a random pick among the candidates only recovers the single-sample reliability.
To evaluate the effect of the verifier, we compare \ours end to end with two variants on the 120 episodes of the six original RoboDojo tasks (\cref{fig:ablation}).
The model, number of samples , initial states, controller, and simulation environments are fixed.
\emph{No verification} executes a candidate drawn uniformly at random from the valid ones.
\emph{No geometric check} removes the geometric check table (principles 1 and 2) and the instruction that refers to it; the verifier still sees all candidates and images and still writes the subgoal note.
No verification lowers success from 44.2\% to 20.0\%, on all six tasks, close to a single sample (17.5\%) as \cref{eq:qk} predicts.
This confirms that the gain of \ours comes from choosing among the candidates, not from drawing more of them.
Without the geometric check table, success falls to 32.5\%, while still 12.5 points better than a random pick.

\begin{figure}[t]
  \centering
  \includegraphics[width=0.6\linewidth]{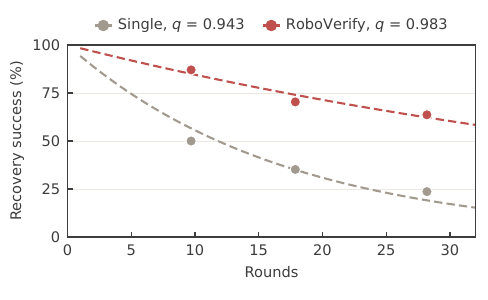}
  \caption{\textbf{Per-round reliability compounds over the horizon.} RoboDojo and PolaRiS episodes grouped into terciles by the number of rounds their recovery takes. Dots: observed success; dashed: $q^n$ with the per-round reliability $q$ fitted for each method.}
  \label{fig:horizon}
\end{figure}

\paragraph{Gains per round compound over the recovery.}
By \cref{eq:compound}, a small gain per round becomes a large gain per task.
A recovery takes many decisions, and in closed loop a wrong one leaves the robot in a state from which later decisions are harder.
This likely explains why the closed-loop error in \cref{fig:scaling} lies above best-of-$k$, most of all with a single sample.
Fitting the equivalent per-round reliability $q$ on RoboDojo and PolaRiS gives $q=0.943$ for \single and $q=0.983$ for \ours.
\ours thus cuts the failure probability per round from 5.7\% to 1.7\%, which over a 20-round recovery is the difference between 31\% and 71\% success.
The success of both methods falls with the horizon as the model predicts (\cref{fig:horizon}).
In short, a single sample of the cheaper model is unreliable, but the correct action is increasingly present among several samples.
Verifying them with grounded evidence raises the reliability of each decision, and because a recovery consists of many decisions, this gain compounds into a large gain in task success.

%% file: sec/7_conclusion.tex
\section{Conclusion}
\label{sec:conclusion}
In this work, we introduce \ours, a model-agnostic, training-free self-verification framework for failure-time takeover: one model samples several candidate actions per decision and verifies them against code-computed geometry and cross-sample measurements.
Our results show that a cheaper model is largely capable of recovering from real policy failures but is unreliable from one sample to the next.
\ours closes most of its gap to a frontier agent on three simulation suites and on a real robot, at 7.4--10.3$\times$ lower cost.
We build \bench, our benchmark of real $\pi_{0.5}$ failures, to support further work on recovery.
Our central message is that, on some tasks, cheaper models can take the place of frontier models in robot control with much lower cost.
Nothing in \ours is specific to cheaper models: it may equally raise the reliability of stronger models, including frontier ones.
And as cheaper models improve with each generation, the same framework may let them match the frontier models of their time.

%% file: sec/8_acknowledgments.tex
\section*{Acknowledgments}
We thank the members of the Stanford Autonomous Systems Lab, Stanford Scaling Intelligence Lab, Stanford Geometric Computation Group, and Stanford Vision and Learning Lab for their constructive feedback and informative discussions.
This work was supported by Google; Google DeepMind; Google Cloud; Stanford HAI; DARPA (HR00112520038, Fallingwater); NSF (24-554, AIMing); NASA ULI; Lightspeed, and Grant AW1134392 (Reasoning in Motion) from the TRI University 3.0 Program.
We also acknowledge the support of IBM and Felicis as members of Stanford HAI's Industry Affiliates Program.
This material is also based upon work supported by the Defense Advanced Research Projects Agency (DARPA) under Agreement Number 00011869.
Any opinions, findings, and conclusions or recommendations expressed in this material are those of the author(s) and do not necessarily reflect the views of the Defense Advanced Research Projects Agency (DARPA).

%% file: sec/X_suppl.tex
\clearpage
\appendix
{\noindent\sffamily\bfseries\LARGE Appendix\par}\vspace{1.5ex}
\suppressfloats[t]   
\renewcommand{\topfraction}{0.9}\renewcommand{\dbltopfraction}{0.9}\renewcommand{\textfraction}{0.07}\renewcommand{\bottomfraction}{0.6}
\renewcommand{\floatpagefraction}{0.8}\renewcommand{\dblfloatpagefraction}{0.8}
\makeatletter\setlength{\@fptop}{0pt}\setlength{\@fpsep}{12pt plus 4pt}\setlength{\@fpbot}{0pt plus 1fil}\setlength{\@dblfptop}{0pt}\setlength{\@dblfpsep}{12pt plus 4pt}\setlength{\@dblfpbot}{0pt plus 1fil}\makeatother

{\small\noindent\textbf{Contents}\par\smallskip
\noindent\begin{tabular*}{\linewidth}{@{}l@{\extracolsep{\fill}}r@{}}
\ref{sec:supp:benchmark}\quad\nameref{sec:supp:benchmark} & \pageref{sec:supp:benchmark} \\
\ref{sec:supp:real}\quad\nameref{sec:supp:real} & \pageref{sec:supp:real} \\
\ref{sec:supp:prompts}\quad\nameref{sec:supp:prompts} & \pageref{sec:supp:prompts} \\
\ref{sec:supp:results}\quad\nameref{sec:supp:results} & \pageref{sec:supp:results} \\
\ref{sec:supp:cost}\quad\nameref{sec:supp:cost} & \pageref{sec:supp:cost} \\
\ref{sec:supp:analysis_details}\quad\nameref{sec:supp:analysis_details} & \pageref{sec:supp:analysis_details} \\
\ref{sec:supp:analyses}\quad\nameref{sec:supp:analyses} & \pageref{sec:supp:analyses} \\
\ref{sec:supp:qualitative}\quad\nameref{sec:supp:qualitative} & \pageref{sec:supp:qualitative} \\
\end{tabular*}\par}

\section{\suppBench}
\label{sec:supp:benchmark}
\paragraph{Source policies.}
All failures come from $\pi_{0.5}$ checkpoints released by the authors of each suite, run without further training (\cref{tab:policies}).

\begin{table*}[t]
  \centering
  \small
  \setlength{\tabcolsep}{4pt}
  \begin{tabular}{@{}p{0.09\linewidth}>{\raggedright\arraybackslash}p{0.40\linewidth}>{\raggedright\arraybackslash}p{0.47\linewidth}@{}}
    \toprule
    Suite & Checkpoint & Source \\
    \midrule
    RoboDojo & Official $\pi_{0.5}$ baseline, step 59{,}999; served through the XPolicyLab $\pi_{0.5}$ adapter~\cite{xpolicylab2026} with config \nolinkurl{pi05_base_aloha_full_sim_arx} &
    Hugging Face dataset \nolinkurl{RoboDojo-Benchmark/RoboDojo}, path \nolinkurl{ckpt/RoboDojo/Pi_05/RoboDojo-sim-arx_x5-joint-0} \\
    \midrule
    RoboCasa & \nolinkurl{pi05_pretrain_human300}, the multitask checkpoint at step 75{,}000 &
    RoboCasa365 release~\cite{nasiriany2026robocasa365}; Hugging Face \nolinkurl{robocasa/robocasa365_checkpoints}, revision \nolinkurl{c484448} \\
    \midrule
    PolaRiS & \nolinkurl{pi05_droid_jointpos_polaris}, the $\pi_{0.5}$-DROID joint-position checkpoint &
    Released with PolaRiS~\cite{jain2025polaris} \\
    \bottomrule
  \end{tabular}
  \caption{\textbf{Source policies} whose failures make up \bench.}
  \label{tab:policies}
\end{table*}

\paragraph{Exact replay and success checks.}
Each episode stores the full action prefix of the source rollout.
Replaying this prefix in the same simulator reproduces the state at the intervention point, which we verify against the recorded object and robot states before any model is called; all methods start from the same verified snapshot, so comparisons are paired.
The intervention point of every episode is set with the help of a human reviewer.

\paragraph{RoboDojo-220.}
The RoboDojo portion contains 220 episodes from 11 tasks, 20 each: build tower, stack blocks, pack objects into a box, put bottles into a bin, fill an egg holder, hang mugs, align blocks, push T, stack blocks by colour, arrange digits, and press buttons by number.
RoboDojo~\cite{chen2026robodojo} is a unified simulation and real-world manipulation benchmark whose simulation runs in Isaac Sim.
The robot is a pair of ARX X5 arms, each with a parallel-jaw gripper, for bimanual tabletop manipulation.
Three cameras observe the scene: a fixed main camera (cam\_high) and one wrist camera per arm; $\pi_{0.5}$ receives the three RGB streams and a 14-dimensional proprioceptive state (six joints and one gripper per arm).
The agent sees the same three views and reads depth through the measurement tools; for each arm it commands an absolute link6 pose and a continuous gripper opening, which the simulator tracks with inverse kinematics.
Each round executes 10 control steps, and its targets are limited to 5\,cm and 0.35\,rad from the measured pose; the budget is 300 control steps.

\paragraph{RoboCasa-180.}
The RoboCasa portion contains 180 episodes from 18 atomic kitchen tasks, 10 each: close the blender lid, close the fridge, close the toaster-oven door, set up a mug for coffee, navigate the kitchen, open a cabinet, open a drawer, open the stand-mixer head, slide the dishwasher rack, turn off the stove, turn on the electric kettle, the microwave, and the sink faucet, and pick and place from counter to cabinet, counter to stove, drawer to counter, sink to counter, and toaster to counter.
RoboCasa~\cite{nasiriany2024robocasa,nasiriany2026robocasa365} is a large-scale kitchen simulation built on MuJoCo (robosuite), with interactive fixtures such as cabinets, drawers, and appliances, and a wide range of objects.
The robot is PandaOmron: a single Franka Panda arm on an Omron mobile base, with a lifting torso and a binary parallel-jaw gripper.
All three cameras move with the robot: a left and a right agent-view camera and a wrist camera.
The agent commands the end-effector pose in the world frame, which RoboCasa's native operational-space controller tracks; it can also move the base (at most 0.15\,m and 0.4\,rad per round) and the torso (at most 0.1\,m per round).
Control runs at 20\,Hz, each round executes 10 control steps, and the budget is 400 control steps.

\paragraph{PolaRiS-120.}
The PolaRiS portion contains 120 episodes from six tabletop tasks, 20 each: stacking blocks in a kitchen, bussing food into a bowl, moving a latte cup onto a board, putting scissors into a large container, scrubbing a pan with a sponge, and putting tape into a container.
PolaRiS~\cite{jain2025polaris} is a real-to-sim evaluation benchmark: its scenes are reconstructed from real environments, rendered with Gaussian splatting, and simulated in Isaac Sim.
The robot is the DROID platform: a fixed-base Franka Panda arm with a Robotiq 2F-85 parallel-jaw gripper.
Two cameras observe the scene: an external camera and a wrist camera.
The agent commands an end-effector pose and a gripper opening; each round executes 10 control steps, and its targets are limited to 5\,cm and 0.35\,rad from the measured pose.
Control runs at 15\,Hz, and the budget is 300 control steps.

\begin{table*}[t]
  \centering
  \small
  \setlength{\tabcolsep}{4pt}
  \begin{tabular}{@{}>{\raggedright\arraybackslash}p{0.20\linewidth}p{0.76\linewidth}@{}}
    \toprule
    Task & An episode succeeds when \\
    \midrule
    Build tower & The target block rests on its layer of the tower (RoboDojo's support checks for that layer) and is released. \\
    Stack blocks & The released stack is one block taller than at the intervention. \\
    Pack objects into a box & An object that was outside the box at the intervention is inside it and released. \\
    Put bottles into a bin & The target bottle is inside the bin. \\
    Fill an egg holder & An egg is seated in the holder and released. If the episode names a target egg that lies outside the basket and the holder at the intervention, it must be that egg; otherwise any egg from the basket counts. \\
    Hang mugs & The target mug hangs on the rack and is released. \\
    Align blocks & The centres of the three cubes lie within 8\,mm of one straight line in the table plane, and each cube has a face axis within 8\textdegree{} of that line. \\
    Push T & The T block is within 7\,mm and 7\textdegree{} of the target outline, without being lifted. \\
    Stack blocks by colour & The upper block rests on the base block and is released; their centres are within 17.5\,mm and their faces turned at most 30\textdegree{} from each other. \\
    Arrange digits & A digit that was not on its own mat at the intervention is released face up on its own mat. \\
    Press buttons by number & The next unfinished button reaches exactly its required number of presses and is released. \\
    \bottomrule
  \end{tabular}
  \caption{\textbf{Success checks on RoboDojo}, one per task, restricted to the interrupted subgoal.}
  \label{tab:dojochecks}
\end{table*}

\begin{figure}[t]
  \centering
  \includegraphics[width=\linewidth]{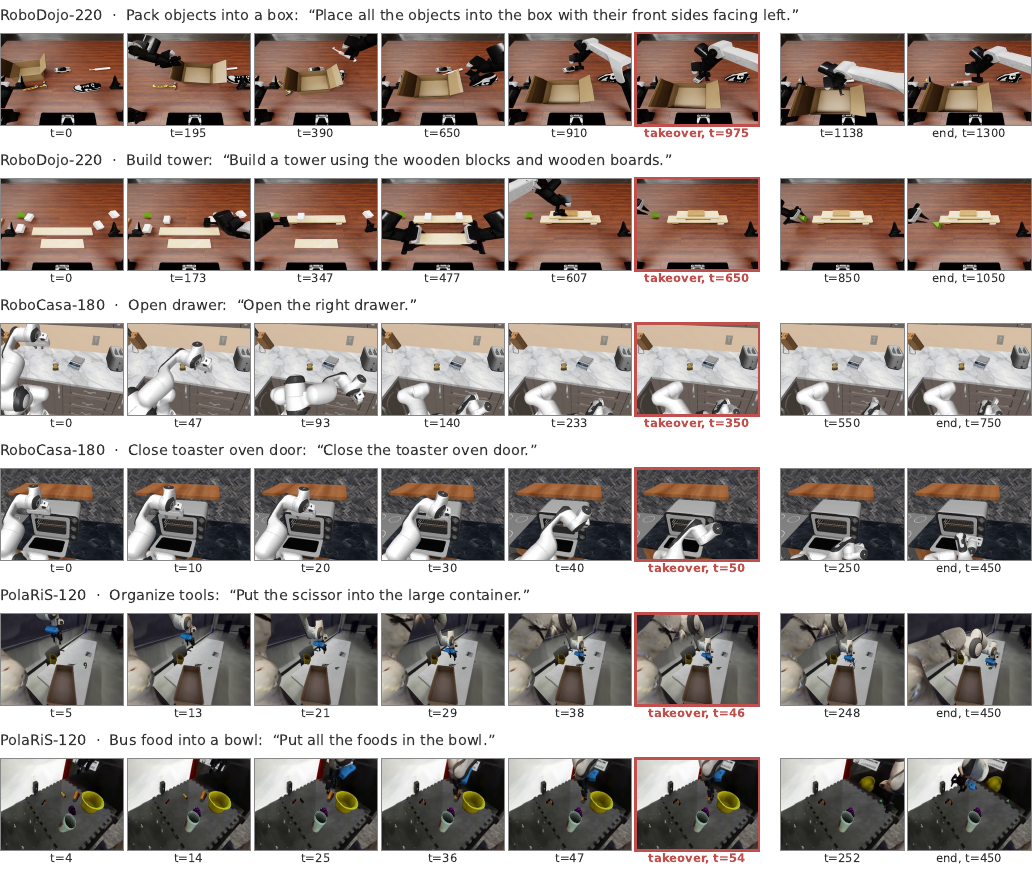}
  \caption{\textbf{Episodes from \bench}, two per suite. Each row is one episode seen from one fixed camera: the source $\pi_{0.5}$ rollout from the start to the takeover point (red frame), where every method takes over, and, on the right, the source policy continuing on its own until its episode ends. $t$ is the control step.}
  \label{fig:simfail}
\end{figure}

\paragraph{RoboDojo success checks.}
After every control step, an episode succeeds when its check (\cref{tab:dojochecks}) holds for the required number of consecutive control steps.
It fails as soon as a RoboDojo task constraint is violated, such as lifting the T block or a wrong or extra button press, and times out after 300 control steps; if RoboDojo itself ends the episode, full task success counts as success and any other end as failure.

\paragraph{RoboCasa success checks.}
RoboCasa has no subgoals: after every control step, RoboCasa's native success predicate of the atomic task checks whether the whole task is complete; once it holds, the episode succeeds and stops, and it times out after 400 control steps.
The native predicates check the scene state by task type: pick and place requires the object in or on its target and the gripper away from the object; doors, drawers, and lids are checked by how far their joints are opened or closed; appliances by their on/off state; and navigation by whether the robot has reached the target fixture.

\begin{table*}[t]
  \centering
  \small
  \setlength{\tabcolsep}{4pt}
  \begin{tabular}{@{}>{\raggedright\arraybackslash}p{0.20\linewidth}p{0.76\linewidth}@{}}
    \toprule
    Task & An episode succeeds when \\
    \midrule
    Stack blocks (kitchen) & A block not yet placed at the intervention is released and rests on the tray: at least 80\% of its horizontal bounding box lies over the tray, and its bottom is within 2.5\,cm of the tray top. If a block already lies on the tray at the intervention, the block must instead rest on it, with at least 50\% over it and its bottom within 2.5\,cm of its top. \\
    Bus food into bowl & A food item not yet in the bowl at the intervention is released and rests in the bowl: at least 80\% of its horizontal bounding box lies over the bowl, and its bottom is below the bowl rim. \\
    Move latte cup to board & The cup is released and rests on the board: in the board's own frame, the centre of the cup's bottom lies within the board, and the cup's lowest point is within 1\,cm of the board top. \\
    Scissors into container & The scissors are released and rest in the large container named in the instruction: at least 80\% of their horizontal bounding box lies over the container, and their bottom is below its rim. \\
    Scrub pan & While the sponge is inside the pan and on its floor (within 7\,cm of the pan centre and 1.6\,cm of the floor), its cumulative horizontal travel reaches 5\,cm. \\
    Tape into container & PolaRiS's native check: the tape is in the container. \\
    \bottomrule
  \end{tabular}
  \caption{\textbf{Success checks on PolaRiS}, one per task.}
  \label{tab:polchecks}
\end{table*}

\paragraph{PolaRiS success checks.}
After every control step, the check of \cref{tab:polchecks} is evaluated; an episode succeeds as soon as it holds, and times out after 300 control steps.
Stacking blocks and bussing food require only the interrupted subgoal; the other four tasks require the whole instruction.
An object counts as released when the Robotiq finger joint is below 0.1\,rad, and as resting when its linear speed is below 0.05\,m/s.

\begin{figure*}[p]
  \centering
  \includegraphics[width=0.82\linewidth]{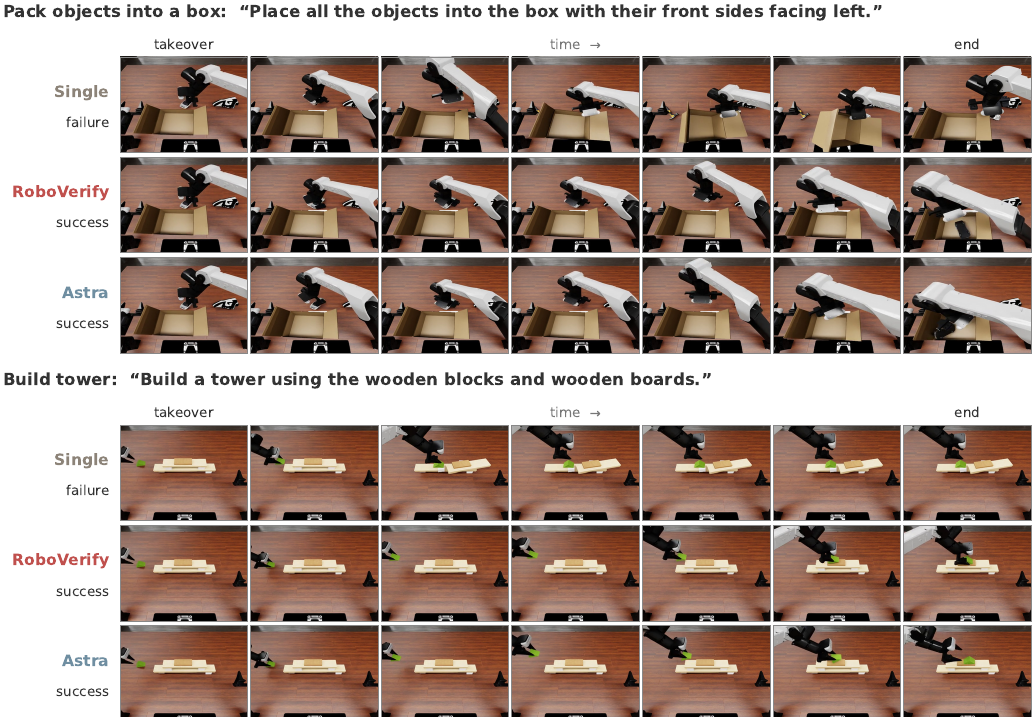}
  \caption{\textbf{Runs on RoboDojo-220} from the takeover states of \cref{fig:simfail}: one run each of \single, \ours, and \astra, from the takeover (left) to the end of the run (right). \ours and \astra complete the interrupted subgoal; \single times out after 300 control steps.}
  \label{fig:simrec:dojo}
\end{figure*}

\begin{figure*}[p]
  \centering
  \includegraphics[width=0.82\linewidth]{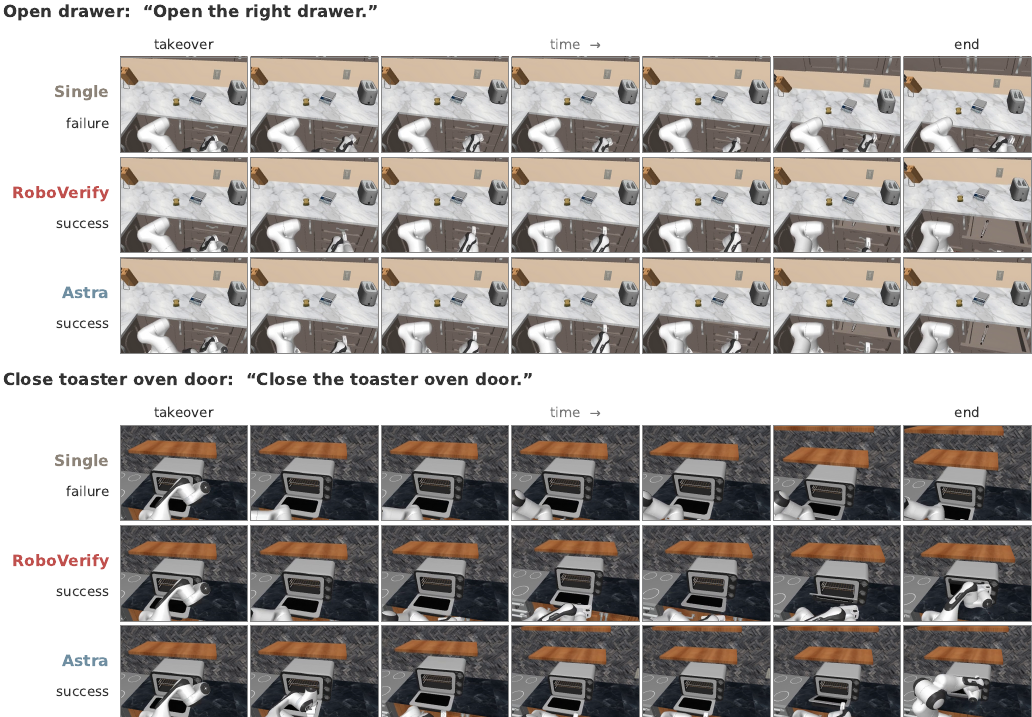}
  \caption{\textbf{Runs on RoboCasa-180} from the takeover states of \cref{fig:simfail}: one run each of \single, \ours, and \astra, from the takeover (left) to the end of the run (right). \ours and \astra complete the task; \single times out after 400 control steps.}
  \label{fig:simrec:casa}
\end{figure*}

\begin{figure*}[p]
  \centering
  \includegraphics[width=0.82\linewidth]{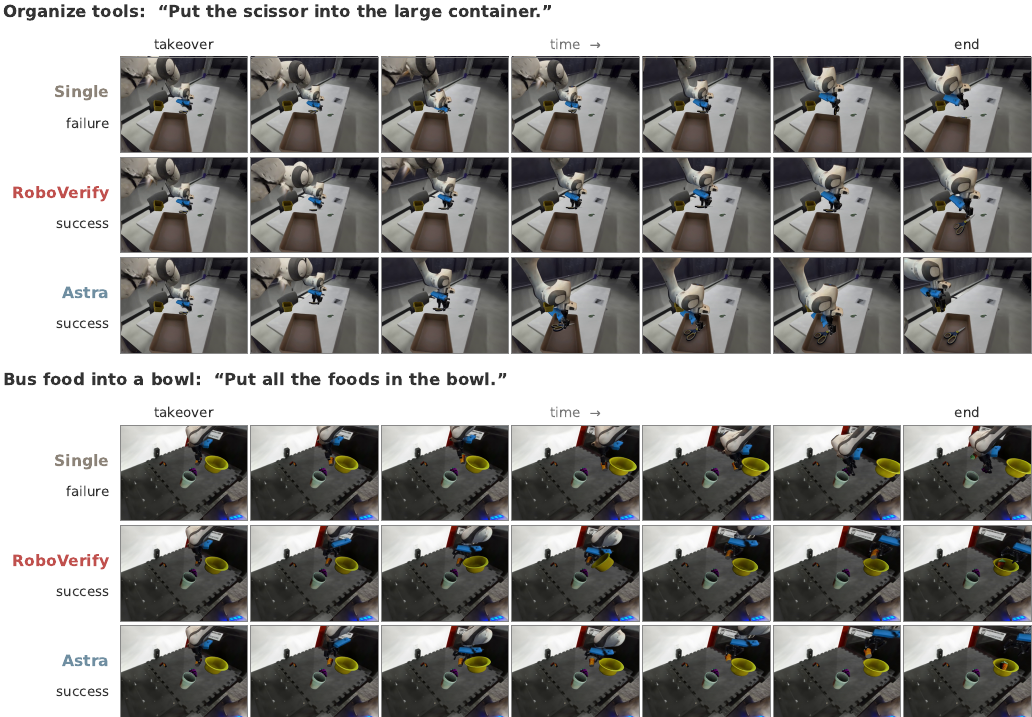}
  \caption{\textbf{Runs on PolaRiS-120} from the takeover states of \cref{fig:simfail}: one run each of \single, \ours, and \astra, from the takeover (left) to the end of the run (right). \ours and \astra complete the interrupted subgoal; \single times out after 300 control steps.}
  \label{fig:simrec:pol}
\end{figure*}

\paragraph{Example episodes.}
\Cref{fig:simfail} shows two episodes from each suite: the source $\pi_{0.5}$ rollout from the start to the takeover point, and how the source policy continues on its own after that point until its episode ends.
\Cref{fig:simrec:dojo,fig:simrec:casa,fig:simrec:pol} show the runs of \single, \ours, and \astra from the same takeover states.

\begin{figure*}[p]
  \centering
  \includegraphics[width=0.82\linewidth]{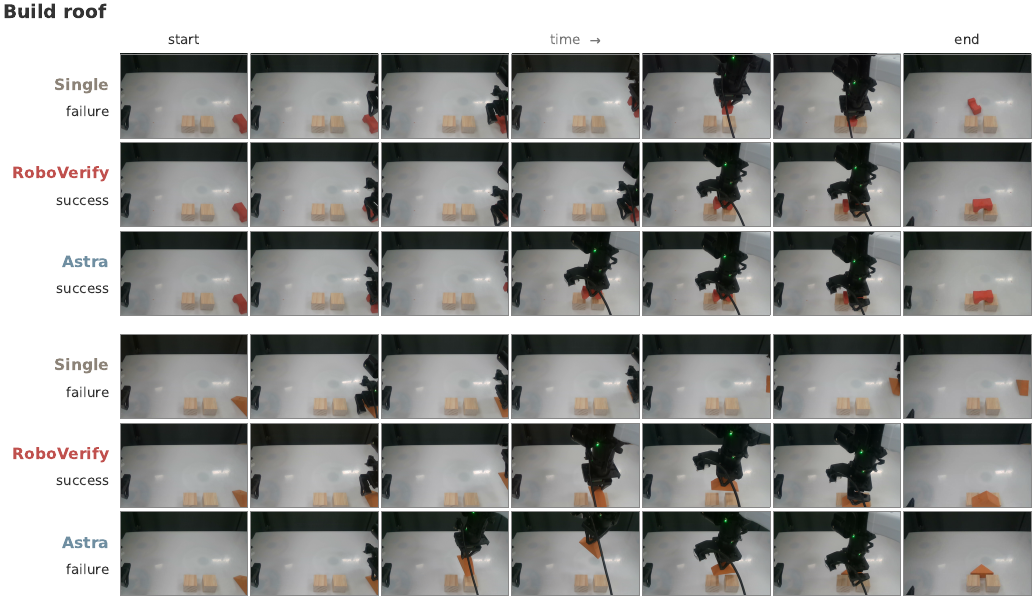}
  \caption{\textbf{Real-robot runs on \emph{Build roof}.} Two scenes (top and bottom); in each, one run of \single, \ours, and \astra from the start (left) to the end of the run (right). \single does not place the roof in either scene; \ours places it in both.}
  \label{fig:realstrips:roof}
\end{figure*}

\begin{figure*}[p]
  \centering
  \includegraphics[width=0.82\linewidth]{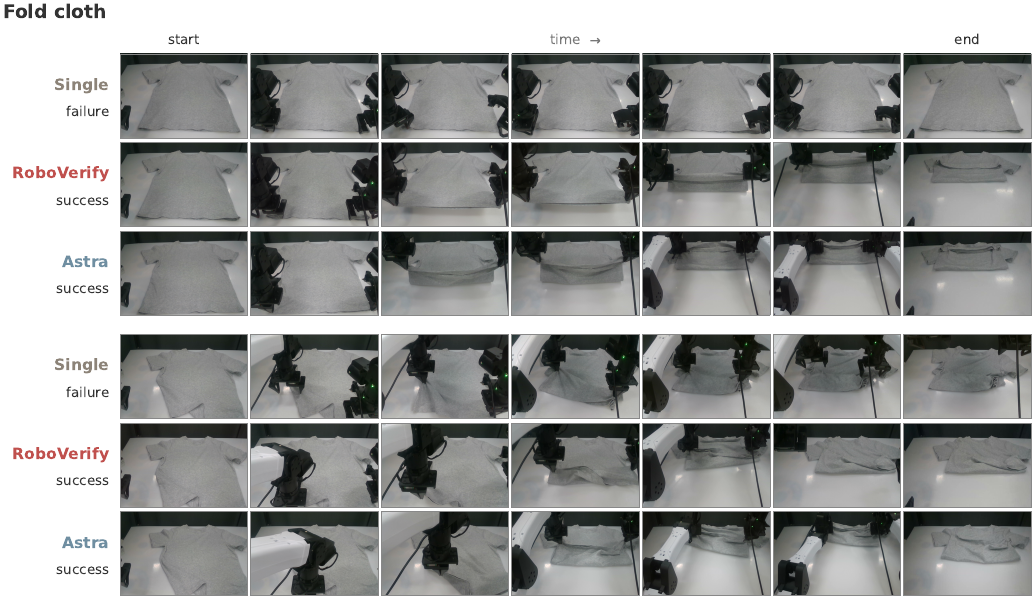}
  \caption{\textbf{Real-robot runs on \emph{Fold cloth}.} Two scenes, a flat shirt (top) and a crumpled shirt (bottom); in each, one run of \single, \ours, and \astra from the start (left) to the end of the run (right). \single does not complete the fold; \ours and \astra fold the shirt in both scenes.}
  \label{fig:realstrips:cloth}
\end{figure*}

\begin{figure*}[p]
  \centering
  \includegraphics[width=0.82\linewidth]{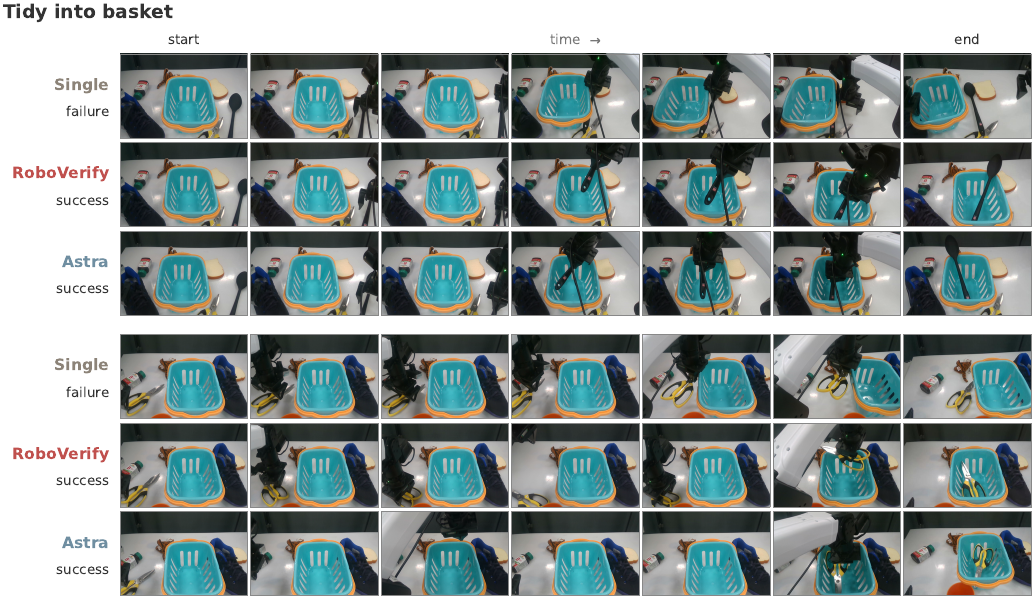}
  \caption{\textbf{Real-robot runs on \emph{Tidy into basket}.} Two scenes, a long spoon (top) and scissors (bottom); in each, one run of \single, \ours, and \astra from the start (left) to the end of the run (right). \single does not put the named object into the basket; \ours and \astra do in both scenes.}
  \label{fig:realstrips:basket}
\end{figure*}

\section{\suppReal}
\label{sec:supp:real}
\paragraph{Hardware and observations.}
Each of the two 6-DoF I2RT YAM arms carries a linear\_4310 parallel-jaw gripper and runs on its own CAN bus.
The agent receives three RGB streams, a fixed overhead camera (cam\_high) and one wrist camera per arm, together with the measured joint states, end-effector poses, and gripper openings.

\paragraph{Action and control.}
The robot executes a 14-dimensional action: six joint targets and one gripper command per arm.
The server on the robot host converts the agent's end-effector targets into joint targets with damped least-squares inverse kinematics on a numerical Jacobian, defined on a virtual tcp\_flange frame.
The control period is 0.04\,s (25\,Hz), and each round of one model decision spans 20 control periods.
Within a round, the translation of the pose target is interpolated by at most 1.5\,cm per period over at most 14 periods, which caps it at 0.21\,m per round; the rotation is interpolated by at most 0.1\,rad per period over at most 17 periods and capped at $\pi/2$\,rad per round; the remaining periods let the arm settle.
An episode lasts at most 300 control periods (12\,s of commanded motion).

\begin{table*}[t]
  \centering
  \small
  \setlength{\tabcolsep}{4pt}
  \begin{tabular}{@{}p{0.11\linewidth}p{0.46\linewidth}p{0.39\linewidth}@{}}
    \toprule
    Task & Instruction & Success criterion \\
    \midrule
    Build roof & Pick up the \emph{[roof block]} and place it on top of the two cubes, so that it forms the roof of the tower. The roof block is a red block or a triangular block. &
    The block rests on the cubes, forming a roof; it stays stable after the gripper has released it and the arm has moved away. \\
    \midrule
    Fold cloth & Fold the cloth once from bottom to top: pick up the two bottom corners of the cloth and fold the bottom edge up, then release both corners and move both arms away from the cloth. \textit{In half of the trials the hem is crumpled or twisted, and the instruction adds that the two bottom corners must first be found in the camera images.} &
    The cloth is folded, and it stays folded after the grippers release it and the arms move away. \\
    \midrule
    Tidy into basket & Pick up the \emph{[object]} and put it into the basket, then release it and move the arm away from the basket. The object is a wooden block, a cup, a long spoon, a pill bottle, scissors, a shoe, or a stuffed toy, among several other objects on the table. &
    The object is inside the basket, not hanging on the rim or lying outside it; it stays in the basket after the gripper releases it and the arm moves away; the basket is not knocked over. \\
    \bottomrule
  \end{tabular}
  \caption{\textbf{Real-robot tasks}: the instruction given to every method and the success criterion used by the operator.}
  \label{tab:realtasks}
\end{table*}

\paragraph{Methods and evaluation.}
\astra, \single, and \ours run against the same server with the same per-round motion limits and task instructions.
All model calls use reasoning effort ``low'' (candidate generation, verification, and \astra).
In each round, every candidate (and \astra) may use at most five model responses, counting the responses that call tools; the fifth must return an action.
Each method runs 8 trials per task, 24 in total.
The trials of the three methods are interleaved: on every task, the methods are run in turn, one trial each, before the next set of trials.
For every trial the operator sets up the scene as described for the task: two cubes side by side in the middle of the table with the roof block lying flat next to them; the shirt within reach of both arms with its hem toward the robot bases; or the named object and an empty basket apart on the table, among other objects.
\Cref{tab:realtasks} gives the success criterion of each task.

\paragraph{Example runs.}
\Cref{fig:realstrips:roof,fig:realstrips:cloth,fig:realstrips:basket} show two scenes per task.
For each scene, the three rows are one run each of \single, \ours, and \astra, seen from the fixed overhead camera from the first round (left) to the end of the run (right).

\section{Prompts and tools}
\label{sec:supp:prompts}
We give the two system prompts of \ours on RoboDojo; the RoboCasa and PolaRiS versions follow the same structure, adapted to their robots and control settings.

\paragraph{Generation prompt.}
Each of the $K$ candidates receives the system prompt below.
In the first round, the user message contains a compact factual history of the source rollout (robot and gripper states every 60 control steps, and every 5 steps over the last 60), the last two pre-failure frames of the three cameras, and the current observation: the original task instruction, the end-effector poses, the measured gripper openings, the public robot geometry, the remaining control budget, and the three current camera views.
Every later round appends the execution outcome and the new observation, preceded by a \texttt{ROUND CONTEXT} block with the remaining controls and the verifier's note (subgoal and stage) from the previous round.
\begin{lstlisting}
Infer the interrupted operation and the recovery subgoal from the original task, pre-failure public history, and current observation. Choose the relevant objects, desired recovered state, and execution plan yourself, and verify your judgement using actual execution feedback. Your proposed objects, subgoals and repair conditions are hypotheses, not environment-provided answers. Action execution success or gripper closure does not establish recovery-subgoal completion. A round may make partial progress. Use public sensor tools to measure rather than guess coordinates. For a new grasp choose a visible region and inspect propose_grasps or propose_point_grasps evidence. Describe each arm role and use rigid_pair_target when moving both holding hands together. Prefer actions that make concrete progress; use a small probing action only when a specific measurement cannot be obtained with tools.
Each observation includes public_robot_geometry derived from measured robot poses: nominal pad centre, approach/jaw axes and their camera projections. Robot interface: absolute link6 position in environment metres, unit quaternion [w,x,y,z]. Robot-local nominal pad-centre offset [0.1501,0,0] metres, verified against the robot asset. Use robot_grasp_geometry and pose_from_contact_axes to inspect grasp axes and compute poses; these tools do not choose where to grasp. A visible surface is not necessarily the desired centre between the pads. Both arms execute for 10 controls, each tracking one pose directly; multi-stage arm motion requires successive segments. Per arm, gripper_timing="with_motion" applies the jaw command immediately; "after_pose" holds its previous jaw command until the requested pose is reached. Choose the timing that implements your intended sequence. "before_motion" applies the jaw command while holding the starting arm pose, then starts moving after measured jaw travel reaches the target or settles; the receipt reports which occurred. It does not establish a grasp. Use it when closure must precede movement. The next observation reports whether the jaw command actually started. Gripper obstruction alone does not establish a retained grasp; compare the observed object motion with the intended effect. Each next target is within 0.05 metres and 0.35 radians of its measured pose. gripper_opening is a continuous command from 0 closed to 1 open; maintaining a holding hand can retain its previous commanded opening. measured_grippers reports actual finger travel separately. Historical gripper slots are commands. Pixel tools use original image dimensions and return visible surface points, not object centres.
Success means the interrupted subgoal is fully completed: the object at its final place, released, and stable for 10 controls, within the remaining control budget; finish with a release and a small retreat that keeps it stable. Contact with other objects matters only if it prevents completing the subgoal.
Your reasoning is not guaranteed to carry over to the next round; your visible output is. Therefore write "state" as the inferred recovery subgoal (which object, its required final placement), the current stage, and the measured evidence for that stage; and write "strategy" as the remaining plan as ordered stages with a completion check for each, plus one fallback if the current stage fails. Keep the same subgoal and plan across rounds unless measured feedback shows it is wrong; if you change it, say why in "reason". A ROUND CONTEXT block, when present, gives the remaining controls and an independent reviewer's note on the subgoal and stage; treat that note as a hypothesis to verify, not as environment truth.
Return exactly one JSON action after any tool calls:
{"mode":"action","next_controller":"model","handoff_goal":null,"state":"observed current state","strategy":"recovery approach","reason":"expected effect and next check","repair_condition":"observable condition for resolving the error","repair_review":{"status":"unresolved","evidence":"current observed evidence"},"steps":10,"targets":{"left":{"position":[0,0,0],"quaternion_wxyz":[1,0,0,0],"gripper_opening":0.5,"gripper_timing":"with_motion"},"right":{"position":[0,0,0],"quaternion_wxyz":[1,0,0,0],"gripper_opening":0.5,"gripper_timing":"with_motion"}}}
Supply actual chosen coordinates. Keep next_controller="model" and handoff_goal=null; there is no later policy handoff. repair_condition and repair_review are your own hypotheses about the interrupted subtask, to be verified against public observations. Do not assume that gripping or lifting alone completes recovery, and do not pursue unrelated remaining parts of the original task. Continue using actual feedback until the evaluator stops the trial. Maximum 300 recovery controls from the initial failed state; up to 10 controls per round. Do not execute or request controls beyond this limit.
\end{lstlisting}

\paragraph{Verifier prompt.}
The verifier receives the system prompt below and a single user message.
Its cached prefix holds the original task with the pre-failure history and frames, followed by one \texttt{ROUND RECORD} per executed round (the state at decision time, the executed candidate's visible output and targets, and the verifier's note).
The current round adds the remaining controls and the previous note, the current observation and camera views, the candidates in fixed index order (each candidate's visible output, tool calls with their results, parsed targets, and motion receipt, without its reasoning), and the geometric check table.
A reply that cannot be parsed is returned with the parsing error, up to three calls per round.
\begin{lstlisting}
You select one of the candidate actions for a robot recovering an interrupted manipulation subtask. Only select; never rewrite an action.
Success means the interrupted subgoal is fully completed: the object at its final place, released, and stable for 10 controls, within the remaining control budget. Choose the candidate that most increases that probability; slow or idle progress is a real cost.
Procedure:
1. State the interrupted subgoal (which object, where it must end up) and the current stage, using the original task, public history and images. If a PREVIOUS SELECTION NOTE is given, keep that subgoal unless current evidence clearly contradicts it.
2. For each candidate, judge it against measured evidence and the GEOMETRIC CHECK table, not against its own explanation: does it act on that object; do commanded targets match measured object and gripper poses; is gripper timing right; is there collision or drop risk; does it make concrete progress within these 10 controls.
3. Candidate explanations are unverified claims. Penalize any candidate whose numbers contradict its claims.
4. Prefer the candidate that continues the plan in progress, unless that plan has visibly failed. When progress is possible, do not choose candidates that only hover or re-observe.
5. Contact with other objects matters only if it prevents completing the subgoal.
Return only JSON:
{"subgoal": str, "stage": str, "per_candidate": [{"index": int, "verdict": "one short line: progress / risk"}], "selected_index": int, "reason": str}
\end{lstlisting}

\paragraph{Real-robot prompts.}
On the real robot, the agent performs whole tasks rather than taking over a failure, so the prompts ask it to complete the original instruction from the start.
\single and every candidate of \ours receive the generation prompt below.
\begin{lstlisting}
Perform the whole task in the original instruction, starting from the current observation; any robot history recorded before the start is context only. The trial is checked after every control and ends as soon as the whole task is complete (where required, with objects released and stable). Therefore, whenever you are asked for another action, the task is not yet confirmed complete: if you have just placed and released the last object in its final place, keep it undisturbed (a small retreat at most); otherwise find what is still unfinished and act on it, and never hold still or merely declare completion. Choose the relevant objects, the order of steps, and the execution plan yourself, and verify your judgement using actual execution feedback. Your proposed objects, steps and completion conditions are hypotheses, not environment-provided answers. Action execution success or gripper closure does not establish completion of a step or of the task. Preserve completed work and avoid unnecessary disturbance. A round may make partial progress. Use public sensor tools to measure rather than guess coordinates. For a new grasp choose a visible region and inspect propose_grasps or propose_point_grasps evidence. Describe each arm role and use rigid_pair_target when moving both holding hands together. If uncertain choose a small action that reveals the state.
Each observation includes public_robot_geometry derived from measured robot poses: nominal pad centre, approach/jaw axes and their camera projections. Robot interface: absolute tcp_flange position in environment metres, unit quaternion [w,x,y,z]. Robot-local nominal pad-centre offset [0.1347,0,0] metres, verified against the robot asset. Use robot_grasp_geometry and pose_from_contact_axes to inspect grasp axes and compute poses; these tools do not choose where to grasp. A visible surface is not necessarily the desired centre between the pads. Both arms execute for 20 controls, each tracking one pose directly; multi-stage arm motion requires successive segments. Per arm, gripper_timing="with_motion" applies the jaw command immediately; "after_pose" holds its previous jaw command until the requested pose is reached. Choose the timing that implements your intended sequence. "before_motion" applies the jaw command while holding the starting arm pose, then starts moving after measured jaw travel reaches the target or settles; the receipt reports which occurred. It does not establish a grasp. Use it when closure must precede movement. The next observation reports whether the jaw command actually started. Gripper obstruction alone does not establish a retained grasp; compare the observed object motion with the intended effect. Each next target is within 0.21 metres and 1.5708 radians of its measured pose; the arm moves there smoothly within the 20 controls. gripper_opening is a continuous command from 0 closed to 1 open; A small measured opening does not mean the jaws closed on nothing: a thin part of an object can be held with the fingers nearly closed. Judge whether an object is between and held by the fingers from the camera images (the wrist view of that hand), not from measured_grippers alone. measured_grippers reports actual finger travel separately. Historical gripper slots are commands. Pixel tools use original image dimensions and return visible surface points, not object centres.
Return exactly one JSON action after any tool calls:
{"mode":"action","next_controller":"model","handoff_goal":null,"state":"observed current state","strategy":"task approach","reason":"expected effect and next check","repair_condition":"observable condition for completing the current step","repair_review":{"status":"unresolved","evidence":"current observed evidence"},"steps":20,"targets":{"left":{"position":[0,0,0],"quaternion_wxyz":[1,0,0,0],"gripper_opening":0.5,"gripper_timing":"with_motion"},"right":{"position":[0,0,0],"quaternion_wxyz":[1,0,0,0],"gripper_opening":0.5,"gripper_timing":"with_motion"}}}
Supply actual chosen coordinates. Keep next_controller="model" and handoff_goal=null; there is no later policy handoff. repair_condition and repair_review are your own hypotheses about completing the current step of the task, to be verified against public observations. Do not assume that gripping or lifting alone completes a step. Continue using actual feedback until the trial is stopped. Maximum 300 controls from the start; up to 20 controls per round; remaining_recovery_controls in the observation is the remaining control budget. Do not execute or request controls beyond this limit.
IMPORTANT: look at the camera images first. Object state (where each object is, whether it is held, where it sits in the gripper, what it rests on) is judged from the camera images; arm and gripper coordinates do not give an object's position. When the images and the numbers disagree, trust the images. In your state description, say what the images show about the object you act on.
Release a held object only after the camera images confirm that it is aligned with its intended final position and orientation and is supported there; if it is not, correct it while still holding it.
Each round allows at most 5 model responses, counting every response that calls tools and the one that returns the action. Use few, decisive tool calls; every tool result reports the calls left this round. The last allowed response has no tools and must return the final JSON action; a round without a valid action by then fails.
\end{lstlisting}
The verifier of \ours receives the system prompt below.
\begin{lstlisting}
You select one of the candidate actions for a robot performing a manipulation task from the start. Only select; never rewrite an action.
Success means the whole task is completed: every object at its final place, released, and stable, within the remaining control budget. Choose the candidate that most increases that probability; slow or idle progress is a real cost.
The trial ends as soon as the task is complete, so while you are asked to select it is not yet confirmed: if the last object has just been placed and released, prefer a candidate that leaves it undisturbed (a small retreat at most); otherwise never prefer a candidate that holds still or only declares completion over one that acts on what is still unfinished.
Procedure:
1. State the current step of the task (which object, where it must end up) and its stage, using the original task, public history and images. If a PREVIOUS SELECTION NOTE is given, keep that step unless current evidence clearly contradicts it.
2. For each candidate, judge it against measured evidence and the GEOMETRIC CHECK table, not against its own explanation: does it act on that object; do commanded targets match measured object and gripper poses; is gripper timing right; is there collision or drop risk; does it make concrete progress within these 20 controls.
3. Candidate explanations are unverified claims. Penalize any candidate whose numbers contradict its claims.
4. Prefer the candidate that continues the plan in progress, unless that plan has visibly failed. When progress is possible, do not choose candidates that only hover or re-observe.
5. Contact with other objects matters only if it prevents completing the task or undoes completed work.
Gripper: A small measured opening does not mean the jaws closed on nothing: a thin part of an object can be held with the fingers nearly closed. Judge whether an object is between and held by the fingers from the camera images (the wrist view of that hand), not from measured_grippers alone.
IMPORTANT: look at the camera images first. Object state (where each object is, whether it is held, where it sits in the gripper, what it rests on) is judged from the camera images; arm and gripper coordinates do not give an object's position. When the images and the numbers disagree, trust the images. In each per-candidate verdict, state what the images show about the object involved.
Release: Release a held object only after the camera images confirm that it is aligned with its intended final position and orientation and is supported there; if it is not, correct it while still holding it.
Return only JSON:
{"subgoal": str, "stage": str, "per_candidate": [{"index": int, "verdict": "one short line: progress / risk"}], "selected_index": int, "reason": str}
\end{lstlisting}

\paragraph{Tools.}
\Cref{tab:tools} lists the tools the agent can call.
All methods use the same tools; \texttt{rigid\_pair\_\allowbreak target} is offered only on the bimanual robots (RoboDojo and the real robot).

\begin{table*}[t]
  \centering
  \small
  \setlength{\tabcolsep}{4pt}
  \begin{tabular}{@{}p{0.27\linewidth}p{0.30\linewidth}p{0.37\linewidth}@{}}
    \toprule
    Tool & Input & Output \\
    \midrule
    \texttt{measure\_pixels} & camera, labelled pixels & 3D surface point, depth, and local depth range of each pixel \\
    \texttt{vector\_between} & two 3D points & displacement and distance \\
    \texttt{propose\_grasps} & camera, arm, image polygon around the object, number of grasps (at most 5) & grasp poses from Contact-GraspNet~\cite{sundermeyer2021contact}, each with the next bounded motion \\
    \texttt{propose\_point\_grasps} & camera, arm, one pixel on the object, number of grasps (at most 2) & object mask, preview image, and top-down grasp poses \\
    \texttt{pose\_from\_contact\_axes} & arm, contact point, approach and jaw directions (optional) & target end-effector pose and the bounded next pose toward it \\
    \texttt{robot\_grasp\_geometry} & none & per arm: approach and jaw axes, gripper-pad centre and aperture, their pixel projections, measured gripper opening \\
    \texttt{rigid\_pair\_target} & translation, pivot, rotation axis and angle & targets for both arms when they hold one object together \\
    \texttt{read\_execution\_history} & control-step range and stride & recorded pre-failure states and actions \\
    \bottomrule
  \end{tabular}
  \caption{\textbf{Tools} available to the agent in every round.}
  \label{tab:tools}
\end{table*}

\paragraph{Geometric check table.}
For every candidate, the table lists the number of tool calls and of failed tool calls.
For every arm, it lists the commanded gripper-pad target and the current pad position, the displacement and its vertical component, the jaw motion (closing, opening, or hold) with the commanded and measured opening and its timing, the rotation, and the three nearest pooled points, each with its label, the candidate that measured it, the tool, and its distance.
It reports quantities only and never marks a candidate as passing or failing.
Positions are in the environment frame and distances in metres.
The commanded pad target is the commanded end-effector pose with the robot's nominal gripper-pad offset applied.
The pooled points are every 3D position returned by a successful tool call of any candidate in the round, each kept with its label, the candidate, and the tool; the three nearest are ranked by Euclidean distance to the commanded pad target.
The jaw motion is closing or opening when the commanded opening differs from the measured one by more than 0.1 of the full range, and hold otherwise.

\paragraph{Action parsing.}
A reply must contain exactly one action object: plain JSON, a single fenced JSON block, or one complete object after free text; several blocks are rejected.
Each arm's target needs a finite position and a non-zero quaternion, which is normalised.
A target beyond the per-round motion limit is not rejected: the arm moves toward it by at most the limit in this round, and the receipt reports the remaining distance.
The verifier's reply must name a valid candidate index and give a reason.

\paragraph{Model settings.}
\single and the candidates of \ours call DeepSeek-V4.1-Flash with thinking enabled, temperature 1.0, and at most 65{,}536 output tokens; the number of tool calls per round is not limited in simulation.
The five candidates of a round start from copies of the same parent conversation and run in parallel; their tool calls are executed one at a time.
A reply that cannot be parsed into an action is returned to the candidate with the parser's error, up to 9 times; a candidate that still fails is dropped, and if only one candidate remains it is executed without a verification call.
The verifier uses the same model in a fresh conversation, with thinking enabled, reasoning effort ``high'', temperature 1.0, at most 65{,}536 output tokens, and no tools.
\astra runs at reasoning effort ``high''.
Real-robot settings are given in \cref{sec:supp:real}.

\section{\suppResults}
\label{sec:supp:results}
\begin{table*}[t]
  \centering
  \small
  \setlength{\tabcolsep}{4.5pt}
  \begin{tabular}{@{}ll c c ccc@{}}
    \toprule
    & & & & \multicolumn{3}{c}{Cost per episode} \\
    \cmidrule{5-7}
    Suite & Method & Recovery success (\%) $\uparrow$ & Gap closed $\uparrow$ & USD $\downarrow$ & USD per success $\downarrow$ & Tokens (M) \\
    \midrule
    \multirow{3}{*}{RoboDojo} & \single & 25.0 & 0\%   & 0.19  & 0.75  & 20.2  \\
                              & \ours   & 50.5 & 85\%  & 1.09  & 2.16  & 102.9 \\
                              & \astra  & 55.0 & 100\% & 11.18 & 20.32 & 6.8   \\
    \midrule
    \multirow{3}{*}{RoboCasa} & \single & 9.4   & 0\%   & 0.25  & 2.67  & 32.8  \\
                              & \ours   & 33.3 & 83\%  & 1.69  & 5.07  & 205.4 \\
                              & \astra  & 38.3 & 100\% & 13.76 & 35.90 & 8.2   \\
    \midrule
    \multirow{3}{*}{PolaRiS}  & \single & 25.8 & 0\%   & 0.14  & 0.54  & 16.8  \\
                              & \ours   & 56.7 & 80\%  & 0.87  & 1.54  & 87.9  \\
                              & \astra  & 64.2 & 100\% & 6.45  & 10.06 & 4.4   \\
    \bottomrule
  \end{tabular}
  \caption{\textbf{Main results.} Gap closed is the fraction of the \single-to-\astra gap recovered. Cost is the API cost at list prices (\cref{sec:experiments}); on RoboDojo, cost and tokens are computed over the episodes with complete per-round logs; tokens are per episode.}
  \label{tab:main}
\end{table*}

\begin{figure}[t]
  \centering
  \includegraphics[width=0.6\linewidth]{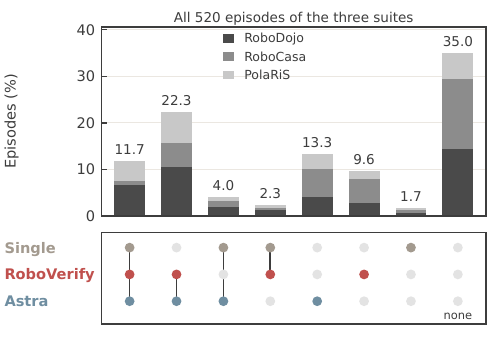}
  \caption{\textbf{Which episodes each method solves}, on all 520 episodes of the three suites. Each column is one combination of methods that succeed (dots below), with its share of all episodes (\%), stacked by suite; the last column counts episodes that no method recovers. Most episodes that \ours recovers are also recovered by \astra.}
  \label{fig:overlap3}
\end{figure}

\paragraph{Which episodes each method solves.}
\cref{fig:overlap3} compares the three methods episode by episode on all 520 episodes.
\single, \ours, and \astra recover 103, 239, and 267 episodes. Of the 239 episodes that \ours recovers, 177 (74\%) are also recovered by \astra.
\ours keeps 73 of the 103 episodes that \single recovers and adds 166 episodes that \single fails, while only 9 episodes (1.7\%) are recovered by \single alone.

\begin{figure*}[t]
  \centering
  \includegraphics[width=\linewidth]{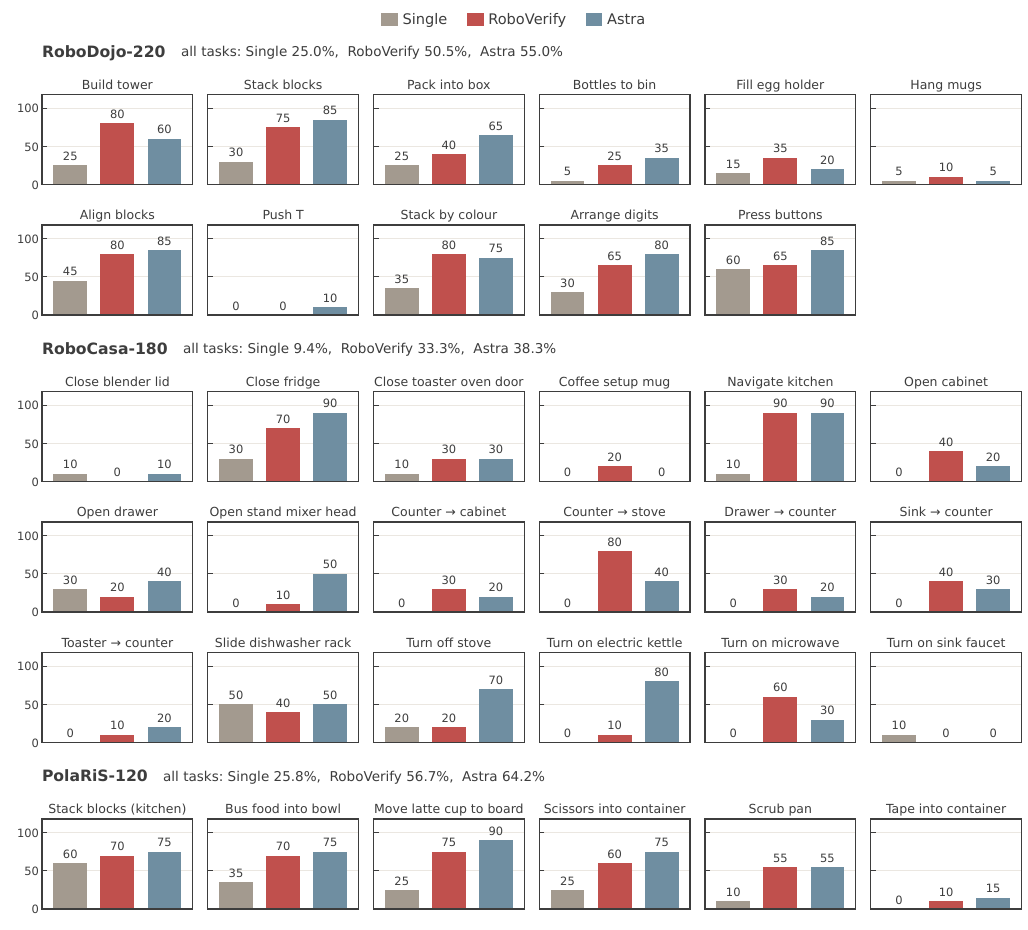}
  \caption{\textbf{Recovery success per task} (\%) on the three suites (RoboDojo and PolaRiS: 20 episodes per task; RoboCasa: 10).}
  \label{fig:pertask_all}
\end{figure*}

\cref{fig:pertask_all} breaks the results down by task on all three suites.

\section{\suppCost}
\label{sec:supp:cost}
On RoboDojo, costs, tokens, and model calls are computed over the episodes with complete per-round logs.
Of the input tokens of \ours, 98.2\%, 99.0\%, and 98.4\% are served from the prompt cache on RoboDojo, RoboCasa, and PolaRiS.
Verification accounts for 16.3\% of the cost of \ours on RoboDojo and 14.9\% on RoboCasa.
Model calls and tokens per method are listed in \cref{tab:tokens}.
On RoboDojo, 764 of the 7{,}371 \astra requests exceed 272K input tokens; on PolaRiS, none do.

\begin{table*}[t]
  \centering
  \small
  \setlength{\tabcolsep}{3.5pt}
  \begin{tabular}{@{}ll r rrr@{}}
    \toprule
    & & & \multicolumn{3}{c}{Tokens per episode} \\
    \cmidrule{4-6}
    Suite & Method & Calls & Input (M) & Cached & Output (M) \\
    \midrule
    \multirow{3}{*}{RoboDojo} & \single & 13{,}934 & 20.0 & 98.6\% & 0.14 \\
                              & \ours   & 71{,}808 & 102.1 & 98.2\% & 0.84 \\
                              & \astra  & 7{,}371 & 6.8 & 96.3\% & 0.011 \\
    \midrule
    \multirow{3}{*}{RoboCasa} & \single & 23{,}006 & 32.7 & 99.1\% & 0.18 \\
                              & \ours   & 139{,}958 & 204.1 & 99.0\% & 1.27 \\
                              & \astra  & 9{,}860 & 8.2 & 96.6\% & 0.013 \\
    \midrule
    \multirow{3}{*}{PolaRiS}  & \single & 12{,}241 & 16.7 & 99.0\% & 0.11 \\
                              & \ours   & 66{,}522 & 87.2 & 98.4\% & 0.69 \\
                              & \astra  & 5{,}149 & 4.3 & 96.0\% & 0.010 \\
    \bottomrule
  \end{tabular}
  \caption{\textbf{Model calls and tokens.} On RoboDojo, calls and tokens are counted over the episodes with complete per-round logs. For \ours, calls include the 5 candidate generations and the verification call of every round (on RoboDojo, 67{,}945 and 3{,}863). Cached is the share of input tokens served from the prompt cache.}
  \label{tab:tokens}
\end{table*}

\section{\suppAnalysis}
\label{sec:supp:analysis_details}

\paragraph{Sampling study (\cref{fig:scaling}).}
We choose RoboDojo failure episodes whose recovery has a single correct target: one object, one arm, and one destination.
A scripted oracle reads the true object poses and bounding boxes from the simulator and computes the next waypoint of a correct recovery.
We use 30 such episodes.
Candidates are drawn from DeepSeek-V4.1-Flash at temperature 1.
In every round the oracle executes the candidate whose commanded gripper target is closest to its next waypoint, with a budget of 200 control steps.
The error of a round is the distance between that target, before any clipping, and the oracle's waypoint, averaged over the rounds of an episode and then over episodes.
Best-of-$k$ takes the closest of the first $k$ of the 16 candidates drawn on the states of the $k{=}16$ run; closed loop is a separate run for each $k\in\{1,3,5,8,12,16\}$.
The error falls from 8.0 to 5.5 and 4.7\,cm ($k{=}1,5,16$) for best-of-$k$, and from 11.3 to 5.8\,cm ($k{=}1$ to 16) in closed loop.

\paragraph{Horizon model (\cref{fig:horizon}).}
We use the 163 episodes that \astra recovers on RoboDojo (86, among the episodes with complete per-round logs) and PolaRiS (77) and take the number of rounds \astra needed as the horizon $n$ of each episode; both suites execute 10 control steps per round.
For each method we fit $q$ by maximum likelihood under $P_{\text{succ}}(n)=q^n$.
Grouping the episodes into terciles of $n$ (54, 54, and 55 episodes), \single succeeds on 50\%, 35\%, and 24\%, against 58\%, 35\%, and 19\% predicted, and \ours on 87\%, 70\%, and 64\%, against 85\%, 74\%, and 62\% predicted.

\section{\suppMore}
\label{sec:supp:analyses}
The analyses below use only logged data; round-level analyses use the 3{,}863 decision rounds of \ours on the RoboDojo episodes with complete per-round logs.

\begin{figure*}[t]
  \centering
  \begin{subfigure}[t]{0.49\linewidth}
    \centering\includegraphics[width=\linewidth]{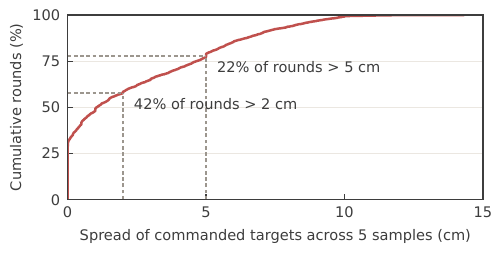}
    \caption{Spread of commanded targets within a round.}
    \label{fig:disagree:cdf}
  \end{subfigure}\hfill
  \begin{subfigure}[t]{0.49\linewidth}
    \centering\includegraphics[width=\linewidth]{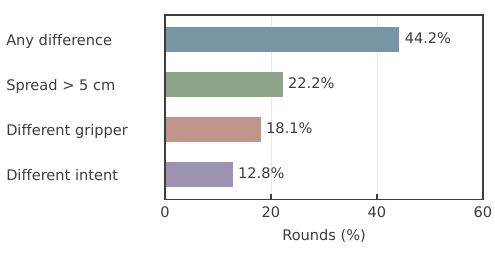}
    \caption{Kinds of disagreement among the five samples.}
    \label{fig:disagree:types}
  \end{subfigure}
  \caption{\textbf{A single sample is unreliable.} In every round, five candidates are sampled from the same state (3{,}863 rounds). (a) Largest distance between the gripper targets they command. (b) Fraction of rounds in which they differ in a substantive way; ``different intent'' means a different arm, a different object near the target, or moving \vs staying still.}
  \label{fig:disagree}
\end{figure*}

\paragraph{Disagreement among samples.}
Candidates differ substantively if their gripper targets lie more than 3\,cm apart, if they command different gripper actions, or if they act with a different arm or on a different object.
\Cref{fig:disagree} details the disagreement summarised in \cref{sec:analysis}: in 44.2\% of rounds the five candidates differ substantively, in 22.2\% their targets lie more than 5\,cm apart, in 18.1\% they command different gripper actions, and in 12.8\% they differ in intent.

\begin{figure*}[t]
  \centering
  \includegraphics[width=0.6\linewidth]{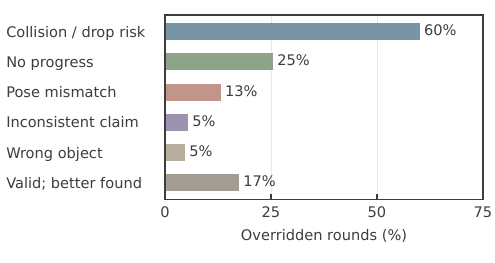}
  \caption{\textbf{What self-verification changes.} Rounds in which the executed action differs substantively from candidate 0, the action a single-sample agent would have executed ($n{=}973$, 25.2\% of rounds), with the verifier's stated reason for rejecting candidate 0 (multi-label, from the verifier's verdicts).}
  \label{fig:select}
\end{figure*}

\paragraph{What self-verification changes.}
We take candidate 0 as the action that \single would have executed.
The verifier picks a different candidate in 68.6\% of rounds; in 25.2\% of rounds the executed action differs substantively from candidate 0 (gripper targets more than 2\,cm apart, or a different gripper action, arm, or object).
In those rounds the verifier's reasons are most often a collision, drop, or constraint risk (60\%), no progress (25\%), or a commanded pose that does not match the measured scene (13\%); in 17\% candidate 0 was judged valid but another candidate was preferred (\cref{fig:select}).

\section{Qualitative examples}
\label{sec:supp:qualitative}
\begin{table*}[t]
  \centering
  \small
  \setlength{\tabcolsep}{4pt}
  \begin{tabular}{@{}p{0.10\linewidth}p{0.05\linewidth}p{0.07\linewidth}p{0.72\linewidth}@{}}
    \toprule
    Candidate & $d$ (cm) & Jaw & Verifier's verdict \\
    \midrule
    \multicolumn{4}{@{}p{\linewidth}@{}}{\textbf{Arrange digits.} Subgoal stated by the verifier: \emph{Pick up the yellow digit `5' with the left gripper and place it on the leftmost wooden pad; final arrangement should form 5310 left-to-right.}} \\
    \midrule
    0 & 2.0 & closing & risk: closes before re-centring from the old offset pose, then lifts; likely repeats the failed grasp. \\
    \textbf{1 (executed)} & \textbf{0.6} & \textbf{closing} & \textbf{progress: re-centres pads to within $\sim$5.5\,mm of measured `5' surface, then closes to 0.2 after alignment.} \\
    2 & 1.0 & opening & partial: re-centres with jaws open only; no grasp closure, so wastes the round. \\
    3 & 4.1 & hold & risk/idle: lifts open 4.8\,cm without aligning or closing; no grasp progress. \\
    4 & 4.3 & closing & risk: closes and lifts from the current off-centre pose, essentially repeating the prior failed grasp. \\
    \multicolumn{4}{@{}p{\linewidth}@{}}{Reason: \emph{Candidate 1 directly fixes the measured off-centre error: its pad target is closest to the measured digit surface, it moves with jaws open to avoid shoving, then closes after alignment. This is the most likely way to establish a retained grasp before lifting and transporting `5' to the leftmost pad.}} \\
    \midrule
    \multicolumn{4}{@{}p{\linewidth}@{}}{\textbf{Stack blocks.} Subgoal stated by the verifier: \emph{Complete a stable stack of the three textured blocks: keep the black block held, place it on the purple block, then place the white block on top.}} \\
    \midrule
    0 & 3.4 & closing & risk: bridging claim contradicts measured white/purple separation; not a centered stack \\
    \textbf{1 (executed)} & \textbf{3.0} & \textbf{closing} & \textbf{progress: moves held black toward the measured purple centre at safe clearance} \\
    2 & 3.2 & closing & risk: aims toward white but target is near the back edge and less certain \\
    3 & 3.2 & closing & risk: y target overshoots past purple centre and remains high \\
    4 & 2.1 & closing & risk: lowers before centering; contact point is off-centre on purple \\
    \multicolumn{4}{@{}p{\linewidth}@{}}{Reason: \emph{The black block is retained and candidate 1 continues the carry toward the clearly measured purple block at constant height, matching the desired purple-centre contact [0.033,$-$0.243] while preserving clearance for the subsequent vertical set-down.}} \\
    \midrule
    \multicolumn{4}{@{}p{\linewidth}@{}}{\textbf{Fill an egg holder.} Subgoal stated by the verifier: \emph{Place the dropped table egg into an empty cup of the egg holder, release it, and leave it stable.}} \\
    \midrule
    0 & 2.7 & hold & Risk: relaxing jaw to 0.55 loses clamp; lift likely drops egg. \\
    1 & 2.7 & closing & Progress: holds 0.30 and lifts; blocked-jaw evidence supports grip. \\
    \textbf{2 (executed)} & \textbf{2.7} & \textbf{closing} & \textbf{Progress: same lift with consistent egg points and asymmetric-contact evidence.} \\
    3 & 2.7 & closing & Progress: same lift, but high-cam points are above the egg and less consistent. \\
    4 & 2.7 & closing & Risk: raising command to 0.43 may loosen the free finger before lift. \\
    \multicolumn{4}{@{}p{\linewidth}@{}}{Reason: \emph{Candidates 1, 2 and 3 command the same 5\,cm lift with jaw held at 0.30, preserving the current clamp; candidate 2 has the most consistent tool evidence that the egg is centered and the jaw is genuinely blocked by it. Candidate 0 relaxes the grip and candidate 4 loosens the closing command, both increasing drop risk.}} \\
    \bottomrule
  \end{tabular}
  \caption{\textbf{Three verifier decisions on RoboDojo.} In each round the verifier rejects candidate~0 and executes the candidate in bold. Its verdicts cite the measured evidence: an off-centre grasp that would repeat a failure, a claim that the measurements contradict, and a command that would loosen the grip on a held egg.}
  \label{tab:qual}
\end{table*}

\paragraph{Verifier decisions.}
\Cref{tab:qual} shows three rounds on RoboDojo in which the verifier executes a different candidate than candidate~0, the action \single would have executed.
For each candidate, $d$ is the distance from the active arm's gripper-pad target to the nearest pooled measured point, and the verdict is the verifier's one-line note; the subgoal and the reason are also the verifier's own words.